# SPINEX: Similarity-based Predictions and Explainable Neighbors Exploration for Regression and Classification Tasks in Machine Learning

M.Z. Naser[1,2], Mohammad Khaled al-Bashiti[1], A.Z. Naser[3]
[1]School of Civil & Environmental Engineering and Earth Sciences (SCEEES), Clemson University, USA
[2]Artificial Intelligence Research Institute for Science and Engineering (AIRISE), Clemson University, USA
E-mail: mznaser@clemson.edu, malbash@g.clemson.edu, Website: www.mznaser.com

[3]Department of Mechanical Engineering, University of Guelph, Canada, E-mail: anaser@uoguelph.ca

## Abstract
The field of machine learning (ML) has witnessed significant advancements in recent years. However, many existing algorithms lack interpretability and struggle with high-dimensional and imbalanced data. This paper proposes SPINEX, a novel similarity-based interpretable neighbor exploration algorithm designed to address these limitations. This algorithm combines ensemble learning and feature interaction analysis to achieve accurate predictions and meaningful insights by quantifying each feature's contribution to predictions and identifying interactions between features, thereby enhancing the interpretability of the algorithm. To evaluate the performance of SPINEX, extensive experiments on 59 synthetic and real datasets were conducted for both regression and classification tasks. The results demonstrate that SPINEX achieves comparative performance and, in some scenarios, may outperform commonly adopted ML algorithms. The same findings demonstrate the effectiveness and competitiveness of SPINEX, making it a promising approach for various real-world applications.

*Keywords:* Algorithm; Machine learning; Interpretability, Supervised learning.

## 1.0 Introduction
The rapid growth of machine learning (ML) techniques has revolutionized various domains, enabling accurate predictions and decision-making [1,2]. In particular, regression and classification algorithms play a pivotal role in extracting valuable insights from data. However, challenges persist, such as the lack of interpretability in complex models [3]. This has led to a growing interest in interpretable ML algorithms [4]. While existing approaches, such as decision trees and linear models, offer interpretability, they often sacrifice predictive performance. Conversely, complex models like neural networks and ensemble methods achieve high accuracy but lack interpretability [5].

In the vast landscape of algorithmic design, similarity-based algorithms are potent methods for tackling various problems [6,7]. These algorithms rely on the premise that similar objects share similar properties and hence draw their strength from leveraging instance similarities and proximity to make predictions or categorizations/clustering. This approach is particularly useful in recommendation systems, classification tasks, and regression problems, where the goal is to predict an outcome based on the similarity of the input data to previously seen examples.

The core idea behind such models is to find the 'neighbors' of a given data point in the feature space. These neighbors are other data points similar to the given instance at hand based on some similarity measure. Once these neighbors are identified, they are used to predict the given data point. This is done by taking a weighted average of the neighbors' outcomes, where the weights are determined by the similarity of each neighbor to the given data point. This underlying principle is harnessed in various contexts, such as document retrieval [8], image recognition [9], etc. The key challenge in realizing such a concept lies in defining "similarity". Traditionally, different similarity measures might be used, such as Euclidean distance, cosine similarity, or the Jaccard coefficient [10].

One of the key advantages of these models is their explainability. Unlike many other ML models, similarity-based models can provide clear explanations for their predictions based on specific, identifiable data points (the neighbors) rather than abstract mathematical functions. However, the concept of 'similarity' is not always straightforward. Different similarity measures may be appropriate for different data types, and choosing the right measure can significantly impact the model's performance. Furthermore, the number of neighbors to consider (the 'neighborhood size') is another important parameter that needs to be carefully chosen [11].

In addition to these challenges, these models also face the 'curse of dimensionality'. This refers to the fact that as the number of features (dimensions) in the data increases, similarity can become less meaningful. This is because, in high-dimensional spaces, all data points tend to be 'far away' from each other, making it difficult to find meaningful neighbors. Despite these challenges, recent research has shown that these models can achieve superior performance on a variety of tasks. For example, these models have been successfully applied to recommendation systems and classification tasks, where they have been shown to outperform traditional collaborative filtering approaches [12,13]. Notable reviews on this front can be found elsewhere [14–16].

One way to implement the similarity-based approach is to use a nearest-neighbor algorithm. The nearest neighbor algorithm finds the *k* most similar objects to a given object and then predicts the label of the given object based on the labels of the *k* nearest neighbors. Such algorithms rely on the idea that the characteristics of an object can be inferred from those of its close neighbors [17]. In a *kNN* algorithm, the class of a new object is determined by the classes of its *k* nearest neighbors in the training set. The "closeness" of objects is typically defined in terms of some distance measure, such as Euclidean distance for numeric data or Hamming distance for categorical data [18]. The strength of such methods is their simplicity and intuitiveness since their predictions are directly tied to the observed data [19,20].

In this context, we propose a novel ML algorithm, SPINEX, which combines similarity-based predictions and neighbors' exploration. SPINEX offers interpretability through feature contribution analysis and interaction effects. Extensive experiments validate the effectiveness and competitiveness of SPINEX in both regression and classification tasks. To evaluate the performance of SPINEX, 59 experiments were conducted on diverse synthetic and real datasets covering a wide range of domains. The experimental results demonstrate the effectiveness and competitiveness of SPINEX compared to state-of-the-art algorithms.

The rest of the paper is organized as follows: Section 2 presents the methodology of SPINEX, explaining each component in detail. Section 3 discusses the experimental setup and presents the results of the experiments conducted. Finally, Section 4 concludes the paper, highlighting the contributions of SPINEX and suggesting potential future directions for research. Therefore, this paper contributes to the field by proposing a novel ML algorithm that combines the strengths of similarity-based predictions and neighbors' exploration, offering interpretability and demonstrating its effectiveness and competitiveness in regression and classification tasks.

## 2.0 Description of SPINEX

This section presents SPINEX via a general description and then a more detailed description.

### *2.1 General description and big ideas*

The SPINEX algorithm comprises several components that provide interpretable regression and classification analysis. First, it begins with a data preprocessing step, which handles missing data, outliers, and feature selection. This preprocessing ensures that the input data is clean and relevant for subsequent analysis. SPINEX then calculates pairwise distances between instances using a user-defined metric. Based on these distances, weights are assigned to the instances using a Gaussian kernel function. The weights reflect the similarity between instances, with closer instances receiving higher weights. This step allows SPINEX to emphasize the influence of the most relevant neighbors during prediction.

SPINEX also accommodates single and ensemble models to capture the inherent complexity of the data. Each model in the ensemble focuses on a specific subset of features, allowing for localized learning and capturing feature interactions more effectively. This algorithm builds on the concept of neighbor-based feature importance, which measures the contribution of each feature to the prediction by considering the influence of its neighboring instances [21]. This approach provides a more nuanced understanding of feature importance, accounting for individual feature effects and their dependencies on nearby instances.

SPINEX also incorporates feature interaction analysis, which explores the interactions between different feature combinations to identify synergistic or antagonistic effects on the target variable. By analyzing the interaction effects, the algorithm unveils the complex relationships among features, shedding light on the underlying mechanisms driving the predictions. By incorporating neighbor exploration, feature contribution, and feature interaction analysis, SPINEX offers insights into the decision-making process, empowering users to make informed decisions based on the predictions.

Further, SPINEX enables the generation of local explanations to gain insights into individual predictions. By considering the neighbors of a specific instance, the algorithm quantifies the importance of each feature and its interaction effects within the local context. The neighbor-based feature importance reveals which features contribute the most to the prediction for the instance at hand, while the interaction effects unveil the synergistic or antagonistic relationships between feature combinations. These local explanations provide a transparent and interpretable view of how the model arrives at its predictions for individual instances.

To facilitate the interpretation and analysis of SPINEX -based models, the proposed algorithm provides various visualization techniques, including feature importance plots, which show the relative contribution of each feature to the prediction, and interaction effect heatmaps, which visualize the interaction effects between different feature combinations. The algorithm also offers tools to analyze the change in predictions with the addition of neighbors, enabling the exploration of the model's behavior in response to nearby instances. These visualizations and analysis methods empower researchers and practitioners to comprehensively understand the model's behavior and uncover the underlying relationships in the data. Table 1 qualitatively compares SPINEX to other commonly used ML algorithms. The same algorithms will be utilized in this work's experiments in a later section.

Table 1 Qualitative comparison between SPINEX and commonly used ML algorithms

| Algorithm | Feature Importance | Interaction Effects | Interpretability | Model Complexity | Ensemble Learning | Handling Categorical Features | Handling Imbalanced Data | Speed | Inference | Model Size | Parallel Computing |
|---|---|---|---|---|---|---|---|---|---|---|---|
| SPINEX | Yes | Yes | Medium | Medium | Yes | No | No | Medium | Fast | Medium | Yes |
| Logistic Regression | Yes | No | High | Low | No | No | No | High | Fast | Small | Yes |
| Decision Tree | Yes | No | High | Varies | No | Yes | No | High | Fast | Varies | Yes |
| Random Forest | Yes | No | Medium | High | Yes | Yes | Yes | Medium | Fast | Large | Yes |
| Gradient Boosting | Yes | No | Medium | High | Yes | Yes | Yes | Low | Fast | Large | Yes |
| AdaBoost | Yes | No | Medium | High | Yes | Yes | Yes | Medium | Fast | Large | Yes |
| CatBoost | Yes | No | Medium | High | Yes | Yes | Yes | Medium | Fast | Large | Yes |
| XGBoost | Yes | No | Medium | High | Yes | Yes | Yes | High | Fast | Large | Yes |
| LightGBM | Yes | No | Medium | High | Yes | Yes | Yes | High | Fast | Large | Yes |
| Support Vector Classifier | No | No | Low | High | No | Yes | Yes | Low | Slow | Large | Yes |
| K-Nearest Neighbors | No | No | High | Low | No | No | No | Low | Slow | Small | Yes |

- Feature Importance: Indicates whether the algorithm can provide feature importance or contribution scores.
- Interaction Effects: Indicates whether the algorithm can capture and quantify interaction effects between features.
- Interpretability: Assesses the ease of understanding and explaining the model's behavior and predictions.
- Model Complexity: Indicates the complexity of the model. It assesses the level of complexity in terms of the number of parameters or rules used by the algorithm.
- Ensemble Learning: Indicates whether the algorithm supports ensemble learning. Ensemble learning combines multiple models to improve performance.
- Handling Categorical Features: Indicates whether the algorithm has built-in mechanisms to handle categorical features.
- Handling Imbalanced Data: Indicates whether the algorithm has techniques to handle imbalanced datasets, where the number of instances in different classes is unequal.
- Speed: Represents the speed of the algorithm for training and prediction tasks. It assesses the algorithm's efficiency in terms of computational time.
- Inference: Indicates whether the algorithm supports efficient inference or prediction on new, unseen data after training.
- Model Size: Assesses the memory footprint or storage requirements of the model.
- Parallel Computing: Indicates whether the algorithm can leverage parallel computing capabilities to speed up training or prediction tasks.

*2.2 Detailed description*

More details are presented to articulate the working mechanisms (i.e., functions and methods) of SPINEX. In addition, SPINEXRegressor and SPINEXClassifier are presented as well.

Data Preprocessing: The SPINEX algorithm applies several preprocessing steps using the DataPreprocessor class. Given data matrix $X \in \mathbb{R}^{n \times d}$ and corresponding label vector $y \in \mathbb{R}^n$, preprocess the data to handle missing values, outliers, and perform feature selection. This includes handling missing data (removing or imputing), outlier detection and removal, and feature selection. The feature selection process uses a local search strategy, and the user can specify prioritized features for inclusion.

Distance Calculation: Distances between instances are calculated in the feature space. For a given distance metric d, calculate the distance $D_{ij}$ between every pair of instances $(x_i, x_j)$ as $D_{ij} = d(x_i, x_j)$. These distances are computed using the specified distance metric (Manhattan distance by default) and are then stored in a distance matrix. The user can specify the distance metric according to the problem at hand.

Weight Calculation: Weights are assigned to each of the training instances based on their distances to the test instance. The weights are computed using the Gaussian kernel function, where instances closer to the test instance will have higher weights such that $w_i$ for each instance $x_i$ based on their distances to the test instance $x_t$ as follows: $w_i = \exp( - (D_{it}^2 / (2\sigma^2) ) )$, where σ is the standard deviation of the distances. The standard deviation for the Gaussian kernel is computed as the mean of the distances.

Prediction: Predictions are made by considering the n nearest neighbors to a given test instance. The label $\hat{y}_t$ for a test instance $x_t$ as $\hat{y}_t = \text{argmax_y} \sum w_i * I(y_i = y)$, where I is the indicator function that is 1 if $y_i$ equals y and 0 otherwise, and the sum is over the n nearest neighbors of the test instance. The number of neighbors, n, is a user-defined parameter. The algorithm identifies these nearest neighbors based on the distance matrix. Once the neighbors are identified, they are combined based on the assigned weights to make a prediction.

Feature contribution refers to the individual impact or importance of each feature on the prediction made by SPINEX. The algorithm calculates feature contributions using a neighbor-based approach. Once the neighbors are determined, the algorithm analyzes the difference in predictions between the original instance and its neighbors. This difference reflects the contribution of each feature to the prediction. Specifically, the algorithm compares the prediction made by the model when a particular feature is present in the instance with the prediction when that feature is absent. Mathematically, the contribution $C_k$ of a feature $f_k$ for an instance $x_i$ as $C_k = P(y|x_i) - P(y|x_i\text{^-}f_k)$, where $P(y|x_i)$ is the prediction probability for the instance $x_i$ and $P(y|x_i\text{^-}f_k)$ is the prediction probability for the instance $x_i$ with the k-th feature excluded. Calculate the interaction effect $I_{kl}$ between two features $f_k$ and $f_l$ as $I_{kl} = C_k + C_l - C_{kl}$, where $C_{kl}$ is the change in prediction probability when both features are excluded.

The larger the difference, the more significant the contribution of the feature to the prediction. The feature contribution calculation takes into account the influence of neighboring instances, allowing the algorithm to capture the contextual importance of each feature. This approach provides a more nuanced understanding of feature importance, as it considers both the individual feature effects and their dependencies on nearby instances.

- In regression and classification algorithms, feature contributions are calculated by predicting the output with and without a given feature, then taking the difference. This indicates how much the prediction changes when a feature is removed, i.e., the "contribution" of the feature.
  - ✓ In the regression algorithm, the method compute_contributions() is used to calculate feature contributions. It does this by predicting the output value with and without each feature and taking the difference.
  - ✓ In the classification algorithm, the method predict_contributions() is used to calculate feature contributions. It does this by predicting class probabilities with and without each feature and taking the difference.
- Interaction effects refer to the combined impact of feature combinations on the prediction made by the SPINEX regression model. The algorithm analyzes the interactions between pairs or sets of features to identify synergistic or antagonistic effects that go beyond the individual contributions of the features. The algorithm examines the change in predictions when specific feature combinations are present or absent from calculating interaction effects. It compares the prediction made by the model with a particular feature combination to the prediction when that feature combination is removed. The difference in predictions reflects the interaction effect between the features in the combination. The algorithm considers all possible feature combinations, ranging from pairs to larger sets, to explore the full landscape of interactions. It quantifies the impact of each feature combination on the prediction, providing insights into the synergies or antagonisms between different features.
  - ✓ In the regression algorithm, compute_combination_impact() is used to calculate interaction effects. It does this by predicting the output value with all features, with one feature removed and two removed, and then calculating the interaction effect as described above.
  - ✓ In the classification algorithm, the method predict_contributions() is used to calculate interaction effects. It does this by predicting class probabilities with all features, with one feature removed and two features removed, and then calculating the interaction effect described above.
- Overall, the calculation of feature contributions and interaction effects is very similar in both the SPINEX classification and regression algorithms. The main difference is in the predicted output (class probabilities vs. output value) and the context in which these calculations are used (classification vs. regression).

Feature Importance and Impact Analysis: Feature importance is calculated as the mean absolute contribution of a feature across all instances. For example, the importance $F_k$ of a feature $f_k$ as $F_k$ $= (1/n) \sum |C_k|$. Calculate the impact $I_F$ of a combination of features F as $I_F = P(y|x_i) - P(y|x_i$^-F), where $P(y|x_i^{-F})$ is the prediction probability when all features in F are excluded. The impact of feature combinations is calculated by excluding a combination of features and computing the change in prediction probabilities. The results are sorted by impact, providing insight into which combinations of features are most important.

- Model Ensembling: The SPINEX model can be used as a base classifier in ensemble models. The user can specify an ensemble method, and the SPINEX classifier is then combined with other classifiers (like DecisionTreeClassifier) to make a final prediction. Three ensemble methods are used in the script: Stacking (where the predictions of the base classifiers are used as input to a final classifier), Bagging (where multiple instances of the SPINEX classifier are trained on random subsets of the training data), and Boosting (where multiple instances of the SPINEX classifier are trained sequentially, with each one focusing on the instances that the previous classifiers misclassified). For base classifiers $h_1, ..., h_p$, in the case of Stacking, calculate the final prediction $h(x)$ as $h(x) = g(h_1(x), ..., h_p(x))$, where g is the final classifier. In the case of Bagging, calculate $h(x)$ as $h(x) = \text{majority}(h_1(x), ..., h_p(x))$. In the case of Boosting, calculate $h(x)$ as $h(x) = \text{weighted_majority}(h_1(x), ..., h_p(x))$.

### *2.3 SPINEX for regression*

```
Inputs:
  X_train: Training feature matrix
  y_train: Training target vector
  X_test: Test feature matrix
  distance_metric: Distance metric for calculating pairwise distances
  num_neighbors: Number of nearest neighbors to consider
  kernel_width: Width of the Gaussian kernel

Procedure SPINEX:
  Preprocess(X_train, X_test)
  distances = CalculateDistances(X_train, X_test)
  weights = CalculateWeights(distances)
  predictions = Predict(X_train, y_train, X_test, weights)
  Return predictions

Procedure Preprocess(X_train, X_test):
  # Handle missing data and outliers in X_train and X_test

Procedure CalculateDistances(X_train, X_test):
  distances = empty matrix of size (number of test instances) x (number of training instances)
  For each test_instance in X_test:
    For each train_instance in X_train:
      distances[test_instance][train_instance] = calculateDistance(test_instance, train_instance, distance_metric)
  Return distances

Procedure CalculateWeights(distances):
  weights = empty matrix of size (number of test instances) x (number of training instances)
```

```
 For each test_instance in distances:
   sorted_distances = sort(distances[test_instance])  # Sort distances in ascending order
   kernel_bandwidth = kernel_width * mean(sorted_distances)  # Compute kernel bandwidth
   For i = 0 to num_neighbors - 1:
     weights[test_instance][i] = calculateWeight(sorted_distances[i], kernel_bandwidth)
 Return weights

Procedure Predict(X_train, y_train, X_test, weights):
 predictions = empty vector of size (number of test instances)
 For each test_instance in X_test:
   nearest_neighbors = GetNearestNeighbors(weights[test_instance], num_neighbors)
   prediction = CalculatePrediction(nearest_neighbors, y_train)
   predictions[test_instance] = prediction
 Return predictions

Procedure GetNearestNeighbors(weights, num_neighbors):
 sorted_indices = indices of weights sorted in descending order
 nearest_neighbors = first num_neighbors indices from sorted_indices
 Return nearest_neighbors

Procedure CalculatePrediction(nearest_neighbors, y_train):
 prediction = average of y_train values corresponding to nearest_neighbors
 Return prediction
```

Explanation of main functions and methods:

- Fitting and Predicting with SPINEXRegressor
  - fit(X, y): This method trains the SPINEX regression model using training data X and target values y.
  - predict(X): This method predicts the target values for a given set of input samples X, using the trained SPINEX regression model.
  - predict_contributions(X): This method predicts the contributions of each feature to the target values for a given set of input samples X.
- Analysis & Visualization of Feature Importance and Interaction Effects
  - get_feature_importance(X): This method calculates the feature importances for each feature in X.
  - get_global_interaction_effects(X): This method calculates the average interaction effects for each feature in the dataset.
  - feature_combination_impact_analysis(X): This method analyzes the impact of different combinations of features on the model's predictions.
  - normalize_importances(importances): A utility function for normalizing feature importances.
  - visualize_feature_importances(local_importances, global_importances, feature_names): This method generates a bar plot to compare local and global feature importances.

- visualize_interaction_effects(interaction_effects_df): This method generates a bar plot to show the interaction effects between different features.
- plot_average_interaction_network(avg_interaction_effects, feature_names=None): This function creates a network graph to visualize the interactions between different features.
- plot_contribution_heatmaps(contributions, interaction_effects, feature_names=None): This function creates two heatmaps - one for individual feature contributions and one for pairwise interactions.

▪ Local Explanations & Visualizations

- get_local_explanation(X, instance_to_explain): This method calculates the local feature importances for a specific instance.
- get_local_interaction_effects(X, instance_to_explain): This method calculates the local interaction effects for a specific instance.
- plot_prediction_change(X, y, instance_to_explain): This function visualizes how the prediction changes as each neighbor is added.
- visualize_neighbor_counts(neighbor_counts): This function visualizes the counts of each neighbor in a bar plot.

▪ Influence of Feature Combinations & Local Changes

- plot_feature_pair_influence(X, instance_to_explain, feature_pair, grid_size=20): This method generates a 3D plot to show how changing the values of a pair of features influences the prediction.
- plot_feature_triplet_influence(X, instance_to_explain, feature_triplet, feature_names, grid_size=20): This method generates a 3D scatter plot to show how changing the values of a triplet of features influences the prediction.
- explore_local_changes(X, instance_to_explain, feature_to_explore, grid_size=100, feature_range=None): This method generates a series of predictions by varying the value of a specific feature and keeping all other features constant.
- plot_local_changes(X, instance_to_explain, feature_to_explain, grid_size=100, feature_range=None, feature_name=None): This function visualizes how the prediction changes as the value of a specific feature changes.
- explore_all_local_changes(X, instance_to_explain, grid_size=100, feature_range=None): This function generates a series of predictions by varying the values of all features one by one and keeping all other features constant.
- explore_local_changes_for_pair(X, instance_to_explain, feature_pair, grid_size=100, feature_range=None): This method generates a series of predictions by varying the values of a pair of features and keeping all other features constant.
- explore_local_changes_for_triplet(X, instance_to_explain, feature_triplet, grid_size=10, feature_range=None): This method generates a series of predictions by varying the values of a triplet of features and keeping all other features constant.

The following are the hyperparameters for the regression version of SPINEX (many of which are similar to those for the classification version):

- n_neighbors: This parameter controls the number of neighbors to use for neighbor queries. The choice of n_neighbors affects the predictions made by the model: a smaller number makes the model more sensitive to local variations in the data, while a larger number makes the predictions more stable at the expense of potentially ignoring smaller patterns.
- distance_threshold: This parameter is used in the calculation of instance weights. Weights are calculated as the reciprocal of the sum of the distance to each neighbor and the decayed distance threshold. The distance_threshold parameter thus controls how much influence more distant neighbors have on the prediction of a given instance.
- distance_threshold_decay: This parameter controls the decay rate of the distance threshold. A lower decay rate means that the influence of more distant neighbors decays more quickly.
- ensemble_method: This parameter allows the user to specify an ensemble method to use in combination with SPINEX. Options include bagging, boosting, and stacking. Ensemble methods combine the predictions of multiple models to improve predictive performance.
- n_features_to_select: This parameter controls the number of features to select for training the model. If auto_select_features is set to True, the model will automatically select the features it deems most important.
- auto_select_features: If this parameter is set to True, the model will automatically select a subset of features for training. The number of features to select is controlled by the n_features_to_select parameter.
- use_local_search: If this parameter is set to True, the model will perform a local search to select the best features for training. This can potentially improve the model's performance but may also increase training time.
- prioritized_features: This parameter allows the user to specify a list of features that should be prioritized in the feature selection process.
- missing_data_method: This parameter allows the user to specify the method for handling missing data. Options include mean_imputation, which replaces missing values with the mean of the existing values, and deletion, which removes instances with missing values.
- outlier_handling_method: This parameter allows the user to specify the method for handling outliers. Options include z_score_outlier_handling, which removes instances that have a Z-score greater than 3, and iqr_outlier_handling, which removes instances that fall outside a certain range defined by the interquartile range (IQR).
- exclude_method: This parameter allows the user to specify a method for excluding certain instances from the training set. This could be used, for example, to exclude instances considered outliers based on some criterion.

## *2.4 SPINEX for classification*

```
Inputs:
  X_train: Training feature matrix
  y_train: Training target vector (class labels)
  X_test: Test feature matrix
  distance_metric: Distance metric for calculating pairwise distances
  num_neighbors: Number of nearest neighbors to consider
  kernel_width: Width of the Gaussian kernel

Procedure SPINEX:
  Preprocess(X_train, X_test)
  distances = CalculateDistances(X_train, X_test)
  weights = CalculateWeights(distances)
  predictions = Predict(X_train, y_train, X_test, weights)
  Return predictions

Procedure Preprocess(X_train, X_test):
  # Handle missing data and outliers in X_train and X_test

Procedure CalculateDistances(X_train, X_test):
  distances = empty matrix of size (number of test instances) x (number of training instances)
  For each test_instance in X_test:
    For each train_instance in X_train:
      distances[test_instance][train_instance] = calculateDistance(test_instance, train_instance, distance_metric)
  Return distances

Procedure CalculateWeights(distances):
  weights = empty matrix of size (number of test instances) x (number of training instances)
  For each test_instance in distances:
    sorted_distances = sort(distances[test_instance])  # Sort distances in ascending order
    kernel_bandwidth = kernel_width * mean(sorted_distances)  # Compute kernel bandwidth
    For i = 0 to num_neighbors - 1:
      weights[test_instance][i] = calculateWeight(sorted_distances[i], kernel_bandwidth)
  Return weights

Procedure Predict(X_train, y_train, X_test, weights):
  predictions = empty vector of size (number of test instances)
  For each test_instance in X_test:
    nearest_neighbors = FindNearestNeighbors(weights[test_instance], num_neighbors)
    prediction = CalculatePrediction(nearest_neighbors, y_train)
    predictions[test_instance] = prediction
  Return predictions

Procedure FindNearestNeighbors(weights, num_neighbors):
  sorted_indices = indices of weights sorted in descending order
  nearest_neighbors = first num_neighbors indices from sorted_indices
  Return nearest_neighbors

Procedure CalculatePrediction(nearest_neighbors, y_train):
```

```
class_counts = empty dictionary
For each neighbor in nearest_neighbors:
  class_label = y_train[neighbor]
  If class_label is not in class_counts:
    class_counts[class_label] = 1
  Else:
    class_counts[class_label] += 1
prediction = class label with the highest count in class_counts
Return prediction
```

Explanation of main functions and methods:

- Fitting and Predicting with SPINEXClassifier
  - fit(X, y): This method trains the SPINEX classification model using training data X and target values y.
  - predict(X): This method predicts the target values for a given set of input samples X, using the trained SPINEX classification model.
- Assessing Model Accuracy
  - score(X, y): This method calculates the mean accuracy on the given test data and labels.
- Analysis & Visualization of Feature Importance and Interaction Effects
  - get_feature_contributions(X): This method calculates the contributions of each feature for a given instance.
  - plot_feature_contributions(feature_contributions): This method generates a bar plot to visualize the contributions of each feature.
  - get_feature_interactions(X): This method calculates the interactions between every pair of features for a given instance.
  - plot_feature_interactions(feature_interactions): This method generates a heatmap to visualize the interaction effects between different features.
  - plot_prediction_change(X, y): This function visualizes how the prediction changes as each neighbor is added.
- Local Explanations & Visualizations
  - get_local_explanation(X, instance_to_explain): This method calculates the local feature importances and neighbor counts for a specific instance.
  - get_local_interaction_effects(X, instance_to_explain): This method calculates the local interaction effects for a specific instance.
  - visualize_neighbor_counts(neighbor_counts): This function visualizes the counts of each neighbor in a bar plot.
- Influence of Feature Combinations & Local Changes
  - plot_all_feature_contributions(model, X): This function generates a scatter plot to visualize the contributions of all features.
  - get_global_interaction_effects(X, y) and feature_combination_impact_analysis(X, y): These functions calculate the average interaction effects and the impact of feature combinations for all instances in the dataset.

- get_feature_importance(X, instance_to_explain): This method obtains the feature importances and interaction effects for selected instances.
- plot_feature_pair_influence(X, instance_to_explain, feature_pair): This method generates a scatter plot to show how changing the values of a pair of features influences the prediction.
- plot_feature_triplet_influence(X, instance_to_explain, feature_triplet): This method generates a scatter plot to show how changing the values of a triplet of features influences the prediction.

## 3.0 Description of benchmarking experiments, algorithms, and datasets

This section describes the experimental examination used to benchmark SPINEX. For a start, SPINEX was examined against ten other commonly used ML algorithms, namely, Logistic Regression, Decision Tree, Random Forest, Gradient Boosting, AdaBoost, CatBoost, XGBoost, LightGBM, Support Vector Classifier, and K-Nearest Neighbors. All of these models were used in their default settings[1].

A series of synthetic and real regression and classification datasets were used. Many of these datasets were also recently benchmarked via several ML algorithms [22,23]. Each dataset is checked per the recommendations of recent researchers aimed at measuring data health. Three criteria were selected, and all of these criteria were satisfied,

- Van Smeden et al. [24] require a minimum set of 10 observations per feature.
- Riley et al. [25] suggest a minimum of 23 cases per feature.
- Frank and Todeschini [26] recommend maintaining a minimum ratio of 3 and 5 between the number of observations and features.

A 5-fold cross-validation technique is applied in regression experiments, and in classification experiments, a stratified 10-fold cross-validation technique is applied [27–29]. The performance of the ML models created is evaluated via a number of regression and classification metrics, as listed in Table 2 [30]. For regression problems, the metrics included the mean absolute error (MAE) and coefficient of determination ($R^2$). In general, lower values of MAE and values close to positive unity for $R^2$ are favorable. In addition, the classification metrics include accuracy, logloss error, and the area under the receiver operating characteristic (ROC) curve (AUC). Naturally, higher values of accuracy and AUC and lower values of the logloss metrics are favorable. Finally, two newly-derived functional metrics are also used, predictivity and estimated energy.

[1] The default setting for SPINEX include:

- SPINEXRegressor = SPINEX(n_neighbors=5, distance_threshold=0.05, distance_threshold_decay=0.05, ensemble_method=None, n_features_to_select=None, auto_select_features=False, use_local_search=False, prioritized_features=None, missing_data_method='mean_imputation', outlier_handling_method='z_score_outlier_handling', exclude_method='zero')
- SPINEXClassifier = SPINEX(n_neighbors=5, distance_threshold=0.05, distance_threshold_decay=0.95, ensemble_method=None, metric='euclidean')

Table 2 List of common performance metrics.

| Metric | Formula |
|---|---|
| Regression | |
| Mean Absolute Error (MAE) | $MAE = \frac{\sum_{i=1}^{n}\lvert P_i - A_i\rvert}{n}$ |
| Coefficient of Determination ($R^2$) | $R^2 = 1 - \sum_{i=1}^{n}(P_i - A_i)^2 / \sum_{i=1}^{n}(A_i - A_{mean})^2$ <br> *A: actual measurements, P: predictions, n: number of data points.* |
| Classification | |
| Accuracy | $ACC = \frac{TP + TN}{P + N}$ <br> *P: predictions, N: number of real negatives, TP: number of true positives, TN: number of true negatives.* |
| Logloss error | $LLE = -\sum_{c=1}^{M} A_i log P$ <br> *M: number of classes, c: class label, y: binary indicator (0 or 1) if c is the correct classification for a given observation.* |
| Area under the ROC curve | $AUC = \sum_{i=1}^{N-1} \frac{1}{2}(FP_{i+1} - FP_i)(TP_{i+1} - TP_i)$ <br> *FP: number of false positives, FN: number of false negatives.* |
| Functional metric (from [22]) | |
| Estimated Energy | $MS \times (TT + PT)$ <br> *MS: model size, TT: training time, and PT: prediction time. Smaller values are favorable with a hypothetical minimum value = 1.0 MB × 10 sec = 10 MB.sec.* |

### *3.1 Synthetic datasets*

A collection of functions that generate synthetic data sets, each simulating different types of relationships between features and the target, were used in our experiments. These datasets are useful for testing and evaluating ML models' performance (see Table 3).

#### 3.1.1 Regression experiments

The make_classification function from the sklearn.datasets module was used herein. More specifically, four primary functions are used. These include:

- generate_regression_data function generates random regression data. The underlying equation for this function can be represented as:
  $y = X_1w_1 + X_2w_2 + ... + X_nw_n + b + \varepsilon$
  Here, 'X' represents the input features, w signifies the weights, b is the bias, and ε is a random noise term.

- generate_synthetic_data function generates a dataset where the relationship between the features and the target is a combination of a quadratic function, a sinusoidal function, and a simple multiplication. This is a somewhat complex relationship, which would provide a challenge to many types of machine learning models. The equation for this function is:
  $y = X_0 + X_1^1 + X_2^2 + ... + X_n^n + \varepsilon + outlier_noise$

In this equation, ε represents random noise, while outlier_noise is additional noise added to randomly chosen samples.

- generate_cubic_data function creates a dataset where the relationship between the features and the target is a combination of a cubic function, a squared function, and a simple multiplication. This type of dataset is useful for testing how well a model can handle cubic relationships. Here, the equation is:
  $y = X_0 + X_1^3 + X_2^4 + ... + X_n^{n+2} + \varepsilon + \text{outlier_noise}$
- generate_exponential_data function creates a dataset with an exponential relationship between one of the features and the target, a decaying exponential for another feature, and simple linear relationships for the remaining features. The equation for this function is:
  $y = \exp(X_0) + X_1^1 + X_2^2 + ... + X_n^n + \varepsilon + \text{outlier_noise}$
- generate_step_data function creates synthetic data where the relationship between the first feature and the output is a step function, and subsequent features contribute polynomially to the output. Gaussian noise is added to the output. The underlying equation can be represented as:
  $y = u(X_0 - 0.5) + X_1^1 + X_2^2 + ... + X_n^n + \varepsilon$
  In this equation, u is the unit step function.
- generate_complex_interaction_data function generates synthetic data with complex interactions between the features, including polynomial, sinusoidal, and logarithmic interactions. Here, the equation is:
  $y = X_0^2 + \sin(X_1) \times \log(X_2^2 + 1) + \varepsilon$
- generate_polynomial_data function creates synthetic data where the relationship between the features and the output is defined by high degree polynomials. The underlying equation can be represented as:
  $y = X_0^3 + X_1^4 - X_2^5 + \varepsilon$
- generate_exp_log_data function generates an interaction between an exponential function of the first feature and the natural logarithm of one plus the second feature. The equation for this function is:
  $y = \exp(X_0) \times \log_1 p(X_1) + \varepsilon$
- generate_sin_exp_data function creates synthetic data where the output is an interaction between a sinusoidal function of the first feature and an exponential function of the second feature. Here, the equation is:
  $y = \sin(\pi X_0) \times \exp(X_1) + \varepsilon$
- generate_tan_data function generates a tangent of the first feature, with subsequent features contributing polynomially to the output. The underlying equation can be represented as:
  $y = \tan(X_0) + X_1^1 + X_2^2 + ... + X_n^n + \varepsilon$

These functions accepts parameters such as:

- n_samples for the number of samples
- n_features for the number of features
- n_informative for the number of informative features, i.e., the features that are useful in predicting the target variable. The remaining features (n_features - n_informative) are generated as random noise.
- noise for the standard deviation of the Gaussian noise applied to the output (dependent variable).
- n_targets for the number of regression targets, i.e., the number of dependent variables. By default, this is set to 1, meaning that the generated dataset will have a single target variable.
- n_outliers for the number of outliers.
- bias for the bias term in the underlying linear model.
- shuffle to assign whether or not to shuffle the samples and the features.
- effective_rank for the approximate number of singular vectors required to explain most of the input data by a linear, low rank model. If None, all features are informative. This parameter can be used to introduce collinearity in the data.
- tail_strength for the relative importance of the fat noisy tail of the singular values profile if effective_rank is not None.
- seed to assign seed for the random number generator, to ensure the reproducibility of the results.

In all functions, acombination of parameters was assigned to create 18 datasets (see Table 3). The amount of noise can be increased or decreased by adjusting the noise_scale parameter. In addition, each function adds a specified number of outliers to the target variable. The outliers are randomly selected from the samples and have a larger amount of normally distributed random noise added to them.

Finally, a dictionary named datasets is created to store the generated synthetic datasets. Each dataset has a descriptive name and is generated by one of the previously described functions, with specific parameters. This dictionary allows easy access to each dataset by its name, which is useful when looping through them later to fit and evaluate different ML models.

Table 3 Datasets used in the regression analysis on synthetic data.

| Dataset | Function | n_samples | n_features | n_informative | n_outliers | noise | bias | shuffle | effective_rank | tail_strength |
|---|---|---|---|---|---|---|---|---|---|---|
| Dataset 1 | generate_regression_data | 50 | 5 | 5 | - | 0.0 | 0.0 | - | - | - |
| Dataset 2 | generate_regression_data | 5000 | 4 | 4 | - | 0.1 | 0.0 | - | - | - |
| Dataset 3 | generate_regression_data | 1000 | 6 | 5 | - | 0.0 | 10 | - | - | - |
| Dataset 4 | generate_regression_data | 7000 | 2 | 2 | - | 0.0 | 0.0 | False | - | - |
| Dataset 5 | generate_regression_data | 750 | 8 | 6 | - | 0.0 | 0.0 | - | 5 | - |
| Dataset 6 | generate_regression_data | 800 | 4 | 4 | - | 0.0 | 0.0 | - | - | 0.1 |
| Dataset 7 | generate_regression_data | 1000 | 5 | 3 | - | 0.0 | 10 | - | - | - |
| Dataset 8 | generate_regression_data | 2500 | 3 | 2 | - | 0.0 | 0.0 | False | - | - |
| Dataset 9 | generate_regression_data | 1000 | 4 | 4 | - | 0.9 | 0.0 | - | 10 | - |
| Dataset 10 | generate_step_data | 2000 | 7 | - | - | 0.0 | - | - | - | - |
| Dataset 11 | generate_cubic_data | 1000 | 10 | - | 20 | 0.5 | - | - | - | - |
| Dataset 12 | generate_synthetic_data | 2000 | 6 | - | 200 | 0.8 | - | - | - | - |
| Dataset 13 | generate_exponential_data | 2000 | 5 | - | 40 | 0.8 | - | - | - | - |
| Dataset 14 | generate_tan_data | 750 | 8 | - | - | 0.1 | - | - | - | - |
| Dataset 15 | generate_complex_interaction_data | 500 | 7 | - | - | 0.0 | - | - | - | - |
| Dataset 16 | generate_polynomial_data | 2000 | 5 | - | - | 0.1 | - | - | - | - |
| Dataset 17 | generate_exp_log_data | 1000 | 10 | - | - | 0.5 | - | - | - | - |
| Dataset 18 | generate_sin_exp_data | 3000 | 5 | - | - | 0.0 | - | - | - | - |

.

A systematic approach to rank models based on the selected multiple metrics is followed. In this approach, we calculate the average scores for each model across different metrics, assign ranks to the models for each metric, calculate the sum-based rank across all metrics, and display the ranked models. This enables a comparative analysis of models based on their performance across various metrics. The outcome of this analysis is shown in Table 4 as well as Fig. 1.

It is quite clear that SPINEX and most of its derivatives rank well in terms of accuracy and the bottom half of the total time for training and prediction. The rankings seem to fall in terms of total time (which also affect the ranking for energy). This is due to the algorithm's design to check for feature pairs and interactions.

Table 4 Ranking results of regression experiment on synthetic data

| Model | MAE | $R^2$ | Rank | Total Time | Estimated Energy | Rank |
|---|---|---|---|---|---|---|
| StackingSPINEX | 1 | 1 | 1 | 15 | 16 | 16 |
| CatBoostRegressor | 6 | 2 | 2 | 16 | 14 | 15 |
| BayesianRidge | 3 | 6 | 3 | 4 | 4 | 4 |
| HuberRegressor | 2 | 8 | 4 | 7 | 7 | 7 |
| GradientBoostingRegressor | 7 | 3 | 4 | 12 | 12 | 12 |
| Ridge | 4 | 7 | 5 | 2 | 1 | 1 |
| XGBRegressor | 8 | 5 | 7 | 9 | 11 | 10 |
| RandomForestRegressor | 9 | 4 | 6 | 14 | 15 | 14 |
| LGBMRegressor | 12 | 9 | 7 | 8 | 8 | 8 |
| BaggingSPINEX | 11 | 10 | 8 | 17 | 17 | 17 |
| Lasso | 5 | 17 | 9 | 1 | 2 | 1 |
| SPINEX | 13 | 11 | 10 | 13 | 13 | 13 |
| BoostingSPINEX | 10 | 15 | 11 | 18 | 18 | 18 |
| KNeighborsRegressor | 14 | 12 | 12 | 5 | 5 | 5 |
| AdaBoostRegressor | 15 | 14 | 13 | 10 | 9 | 9 |
| SVR | 18 | 13 | 14 | 11 | 10 | 11 |
| DecisionTreeRegressor | 17 | 16 | 15 | 6 | 6 | 6 |
| ElasticNet | 16 | 18 | 16 | 3 | 3 | 3 |

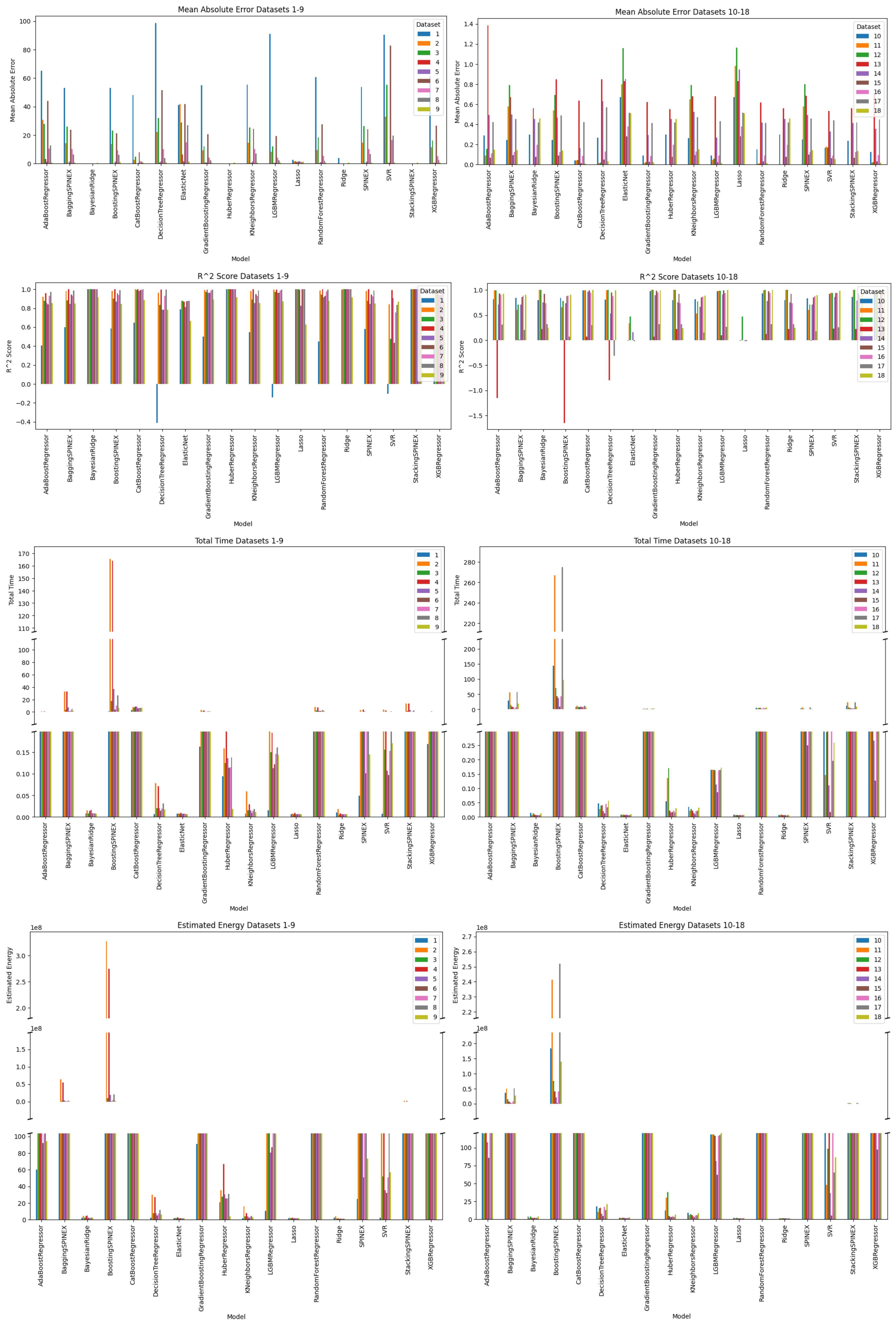

Fig. 1 Regression experiment on synthetic data

3.1.2 Classification experiments

Similar to the regression counterpart, a number of synthetic datasets for binary classification tasks were generated using the make_classification function from the sklearn.datasets module. The function generate_synthetic_data is defined to create synthetic datasets and accepts the following parameters:

- n_samples is the total number of data points in the dataset.
- n_features is the total number of features in the dataset.
- n_informative is the number of informative features (i.e., useful for classifying the samples).
- n_redundant is the number of redundant features (i.e., generated as random linear combinations of the informative features).
- weights is the proportions of samples assigned to each class.
- flip_y is the fraction of samples whose classes are randomly exchanged.
- class_sep is the factor multiplying the hypercube size, wherein larger values spread out the classes tend to make the classification task easier.

The function make_classification generates a random n-class classification problem. It returns X and y, where X is a 2D array of shape n_samples, n_features representing the generated samples, and y is a 1D array of shape n_samples representing the integer labels for class membership of each sample. Overall, 18 datasets were synthetically generated and tested. These were labeled under Series A (see Table 5), and Series B (see Table 6), with varying complexities.

Table 5 Datasets used in the classification analysis on synthetic data (for series A)

| Dataset | n_samples | n_features | n_informative | n_redundant |
|---|---|---|---|---|
| Dataset 1 | 50 | 3 | 2 | 0 |
| Dataset 2 | 100 | 10 | 6 | 2 |
| Dataset 3 | 1000 | 80 | 20 | 40 |
| Dataset 4 | 500 | 20 | 20 | 0 |
| Dataset 5 | 5000 | 40 | 15 | 10 |
| Dataset 6 | 10000 | 10 | 5 | 5 |
| Dataset 7 | 500 | 20 | 20 | 0 |
| Dataset 8 | 3000 | 55 | 20 | 20 |
| Dataset 9 | 50000 | 5 | 3 | 0 |

Table 5 Datasets used in the classification analysis on synthetic data (for series B)

| Dataset | n_samples | n_features | n_informative | n_redundant | flip_y | class_sep | weights |
|---|---|---|---|---|---|---|---|
| Dataset 1 | 50 | 3 | 2 | 0 | 0.01 | 1.0 | 0.9/0.1 |
| Dataset 2 | 100 | 10 | 6 | 2 | 0.02 | 0.5 | 0.8/0.2 |
| Dataset 3 | 1000 | 80 | 20 | 40 | 0.03 | 0.8 | 0.7/0.3 |
| Dataset 4 | 500 | 20 | 20 | 0 | 0.04 | 0.2 | 0.6/0.4 |
| Dataset 5 | 5000 | 40 | 15 | 10 | 0.05 | 0.3 | 0.5/0.5 |
| Dataset 6 | 10000 | 10 | 5 | 5 | 0.06 | 0.4 | 0.6/0.4 |
| Dataset 7 | 1500 | 100 | 40 | 0 | 0.07 | 0.5 | 0.7/0.3 |
| Dataset 8 | 3000 | 55 | 20 | 20 | 0.08 | 0.6 | 0.8/0.2 |
| Dataset 9 | 50000 | 5 | 3 | 0 | 0.09 | 0.7 | 0.6/0.4 |

A similar systematic approach to rank the ML models based on the selected multiple metrics is followed here as that in the regression analysis. The outcome of this analysis is shown in Tables 6 and 7 as well as Figs. 2 and 3. Naturally, most models' predictions are slightly degraded when evaluated on the datasets belonging to Series B (given their complexity).

The outcome of the analysis also shows that SPINEX and its derivatives perform much better in this classification task than in the regression. Overall, and despite its relatively poor ranking under time and energy consumption, the default version of SPINEX consistently ranks in the top 7 in the overall ranking. It is clear that the SPINEX model can outperform some of the more common and traditional algorithms, even in scenarios of imbalanced data and relatively large datasets.

Table 6 Ranking results of classification experiment on synthetic data (Series A)

| Models | Accuracy | LLE | AUC | Rank | Estimated Energy | Total Time | Rank |
|---|---|---|---|---|---|---|---|
| SVC | 1 | 1 | 1 | 1 | 10 | 15 | 13 |
| SPINEXClassifier(default) | 2 | 11 | 2 | 2 | 13 | 9 | 12 |
| StackingSPINEX | 7 | 2 | 7 | 2 | 14 | 12 | 14 |
| BaggingSPINEX | 4 | 7 | 5 | 4 | 15 | 14 | 15 |
| KNeighborsClassifier | 3 | 12 | 3 | 4 | 2 | 2 | 2 |
| BoostingSPINEX | 6 | 5 | 6 | 6 | 11 | 5 | 7 |
| SPINEX | 5 | 6 | 4 | 7 | 12 | 7 | 9 |
| LGBMClassifier | 8 | 4 | 10 | 8 | 4 | 3 | 3 |
| XGBClassifier | 9 | 3 | 9 | 9 | 6 | 6 | 5 |
| RandomForestClassifier | 10 | 8 | 8 | 10 | 8 | 10 | 8 |
| GradientBoostingClassifier | 11 | 9 | 11 | 11 | 9 | 11 | 10 |
| AdaBoostClassifier | 12 | 13 | 12 | 12 | 5 | 8 | 6 |
| DecisionTreeClassifier | 13 | 14 | 14 | 12 | 3 | 4 | 3 |
| LogisticRegression | 14 | 10 | 13 | 14 | 1 | 1 | 1 |
| CatBoostClassifier | - | - | - | - | 7 | 13 | 10 |

*SPINEX = (n_neighbors=20, distance_threshold=0.05, distance_threshold_decay=0.95, ensemble_method=None, metric='manhattan')

Table 7 Ranking results of classification experiment on synthetic data (Series B)

| Models | Accuracy | LLE | AUC | Rank | Estimated Energy | Total Time | Rank |
|---|---|---|---|---|---|---|---|
| SPINEXClassifier(default) | 3 | 1 | 1 | 1 | 13 | 9 | 12 |
| StackingSPINEX | 2 | 3 | 3 | 2 | 14 | 12 | 14 |
| SPINEX | 7 | 2 | 2 | 3 | 12 | 7 | 9 |
| KNeighborsClassifier | 4 | 5 | 5 | 4 | 2 | 2 | 2 |
| BaggingSPINEX | 6 | 4 | 4 | 4 | 15 | 15 | 15 |
| LogisticRegression | 14 | 6 | 6 | 6 | 1 | 1 | 1 |
| RandomForestClassifier | 11 | 8 | 8 | 7 | 9 | 11 | 10 |
| DecisionTreeClassifier | 15 | 7 | 7 | 8 | 3 | 4 | 3 |
| SVC | 1 | 14 | 14 | 8 | 8 | 10 | 8 |
| CatBoostClassifier | 8 | 11 | 11 | 10 | 7 | 14 | 11 |
| GradientBoostingClassifier | 12 | 9 | 9 | 10 | 10 | 13 | 13 |
| AdaBoostClassifier | 13 | 10 | 10 | 12 | 5 | 8 | 6 |
| XGBClassifier | 9 | 12 | 12 | 12 | 6 | 6 | 5 |
| BoostingSPINEX | 5 | 15 | 15 | 14 | 11 | 5 | 7 |
| LGBMClassifier | 10 | 13 | 13 | 15 | 4 | 3 | 3 |

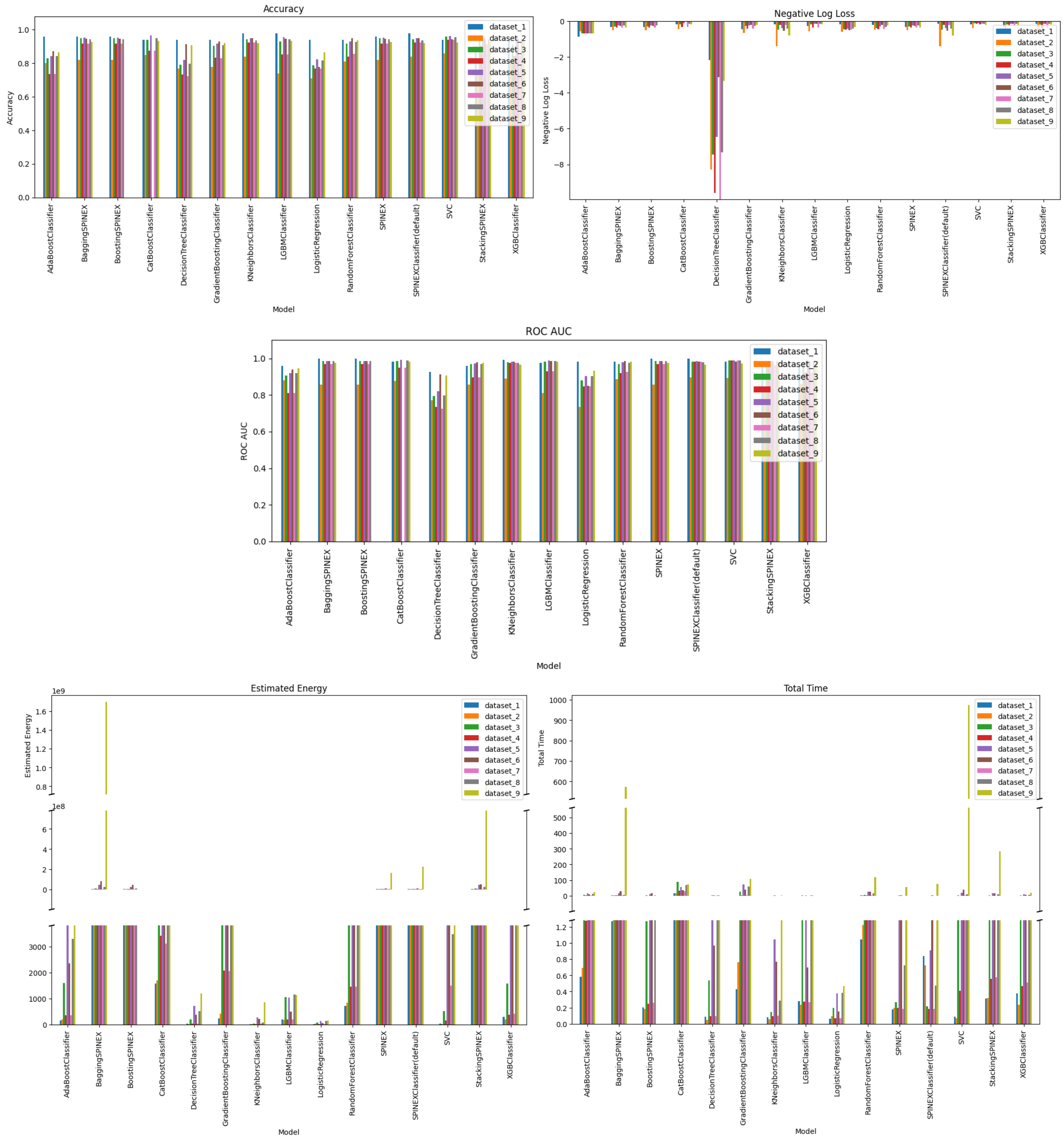

Fig. 2 Classification experiment on synthetic data (Series A)

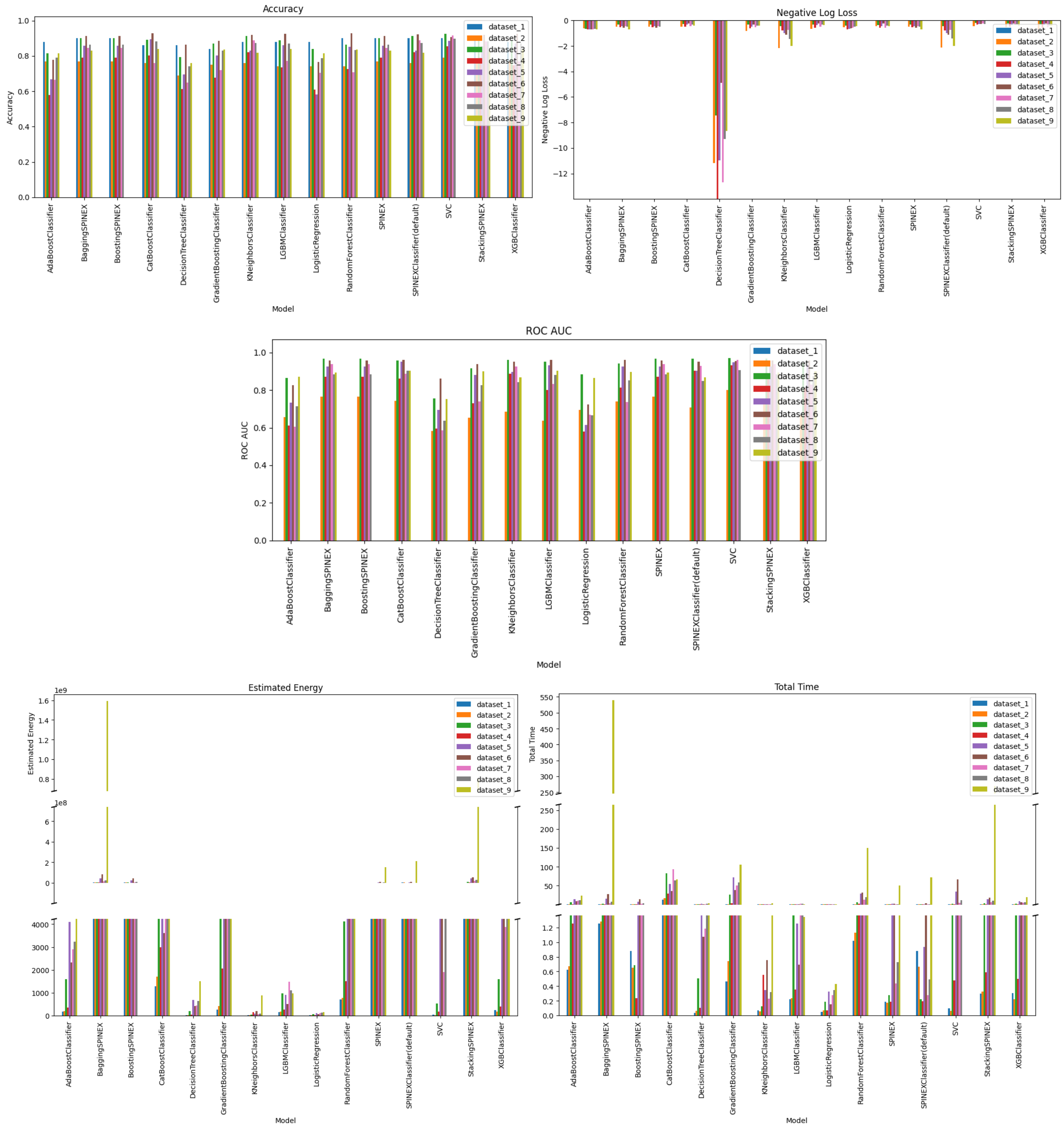

Fig. 3 Classification experiment on synthetic data (Series B)

### *3.2 Real datasets*

Now, we repeat the analysis using a series of real datasets. These datasets are described in the following sections.

#### 3.2.1 Regression experiments

Here, 12 real datasets are used to benchmark SPINEX and the other ML models. Table 7 lists details on each dataset, along with their respective references. As one can see, the selected datasets comprise a collection of samples and features and cover various problem domains. Further information can be found in each dataset's reference.

Table 7 Datasets used in the regression analysis on synthetic data

| Dataset | n_samples | n_features | Ref. |
|---|---|---|---|
| University Admission | 401 | 8 | [31] |
| Fire Resistance of RC columns | 311 | 13 | [32] |
| Shear Strength of beams | 168 | 7 | [23] |
| Concrete Strength | 1031 | 9 | [33] |
| Deformation of Beams under Fire | 1187 | 7 | [34] |
| Strength of Steel Tubes | 1260 | 6 | [35] |
| Energy Efficiency of Buildings | 767 | 10 | [36] |
| Body Fat Index | 252 | 15 | [37] |
| Forest Fire Area | 517 | 13 | [38] |
| Abalone Age | 2000 | 10 | [39] |
| Synchronous Motor | 557 | 5 | [40] |
| Walmart Retail | 3000 | 6 | [41] |

A comparative analysis of models based on their performance across various metrics is presented in Table 8 as well as Fig. 3. The top performing SPINEX derivative ranks 3rd and 4th in terms of accuracy. Other SPINEX derivates also faired well in terms of accuracy metrics (and outperforming some of the common algorithms such as LGBMRegressor, RandomForestRegressor, and GradientBoostingRegressor, but continue to rank at the bottom half due to their large time and energy used.

Table 8 Ranking results of regression experiment on synthetic data

| Model | MAE | $R^2$ | Rank | Total Time | Estimated Energy | Rank |
|---|---|---|---|---|---|---|
| CatBoostRegressor | 1 | 1 | 1 | 3 | 3 | 2 |
| BaggingSPINEX | 8 | 2 | 2 | 14 | 15 | 14 |
| SPINEX | 4 | 7 | 3 | 13 | 14 | 12 |
| LGBMRegressor | 9 | 4 | 4 | 4 | 4 | 3 |
| RandomForestRegressor | 3 | 10 | 5 | 12 | 12 | 11 |
| StackingSPINEX | 11 | 3 | 6 | 16 | 16 | 15 |
| GradientBoostingRegressor | 5 | 12 | 7 | 15 | 13 | 13 |
| HuberRegressor | 12 | 6 | 8 | 6 | 6 | 5 |
| AdaBoostRegressor | 13 | 5 | 9 | 8 | 10 | 8 |
| XGBRegressor | 2 | 16 | 10 | 10 | 11 | 10 |
| KNeighborsRegressor | 10 | 11 | 11 | 5 | 5 | 4 |
| BayesianRidge | 14 | 8 | 12 | 1 | 2 | 1 |
| Ridge | 15 | 9 | 13 | 2 | 1 | 1 |
| DecisionTreeRegressor | 7 | 17 | 14 | 9 | 8 | 7 |
| BoostingSPINEX | 6 | 18 | 15 | 18 | 18 | 17 |
| Lasso | 16 | 14 | 16 | 7 | 7 | 6 |
| SVR | 18 | 13 | 17 | 17 | 17 | 16 |
| ElasticNet | 17 | 15 | 18 | 11 | 9 | 9 |

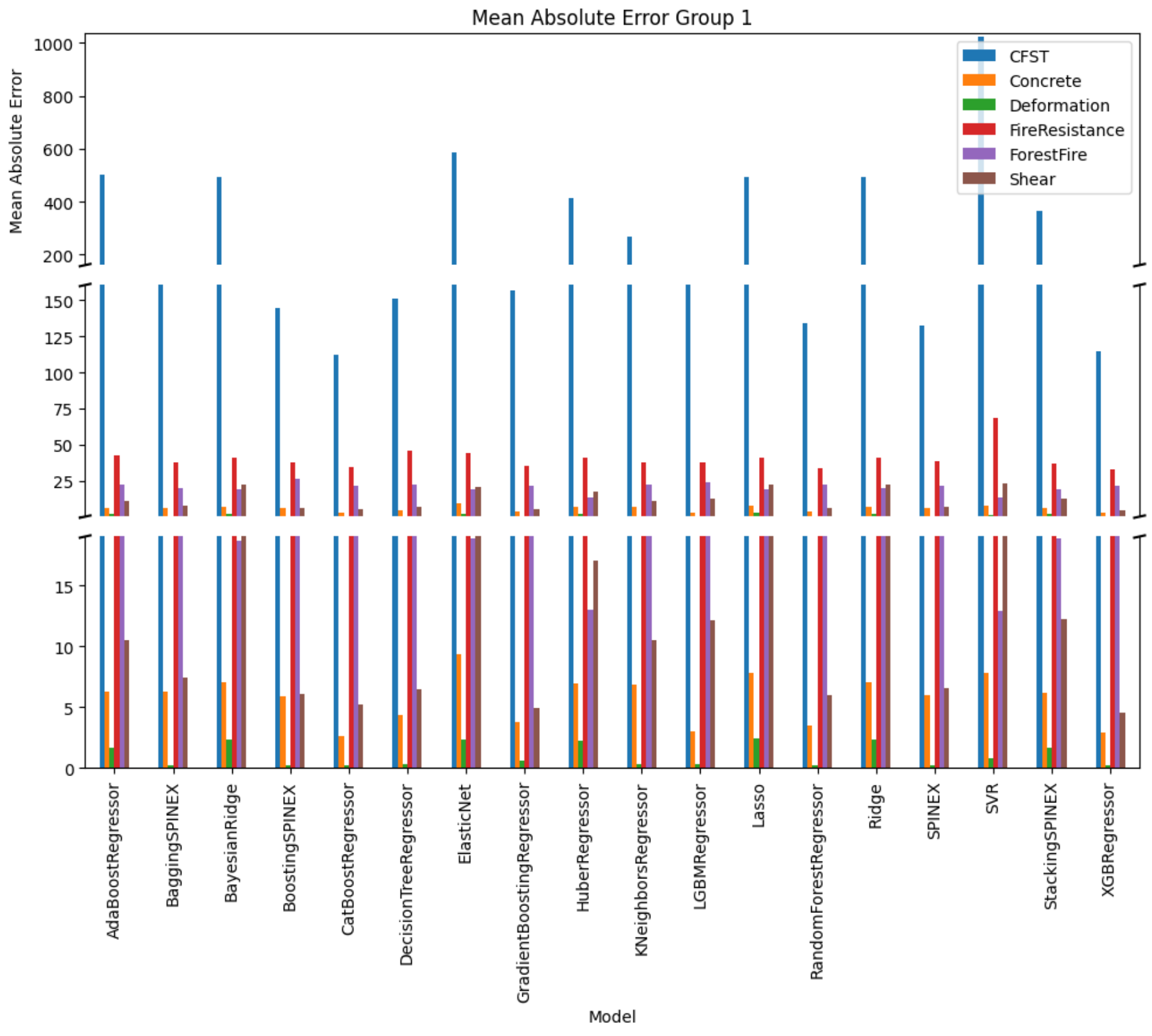

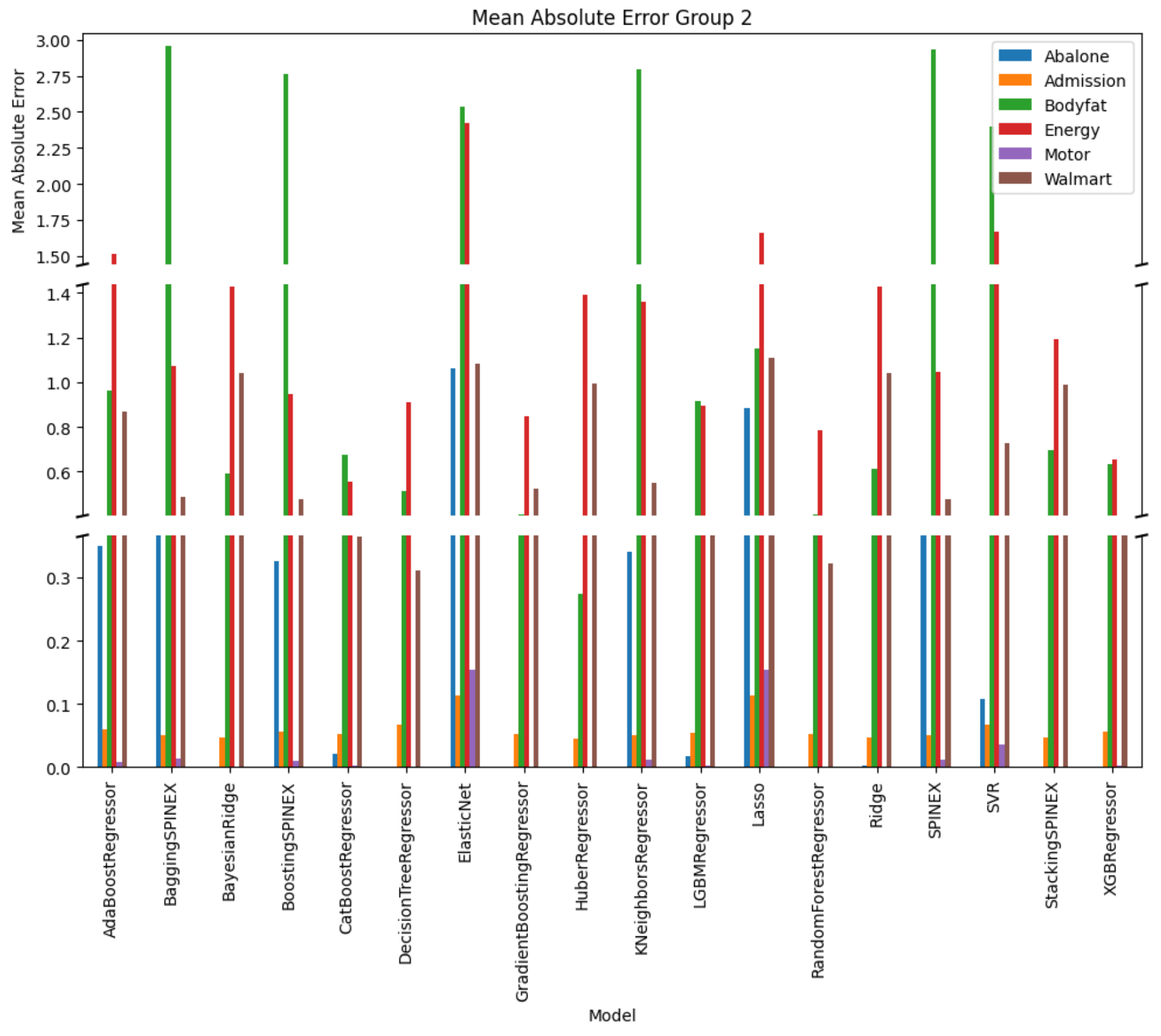

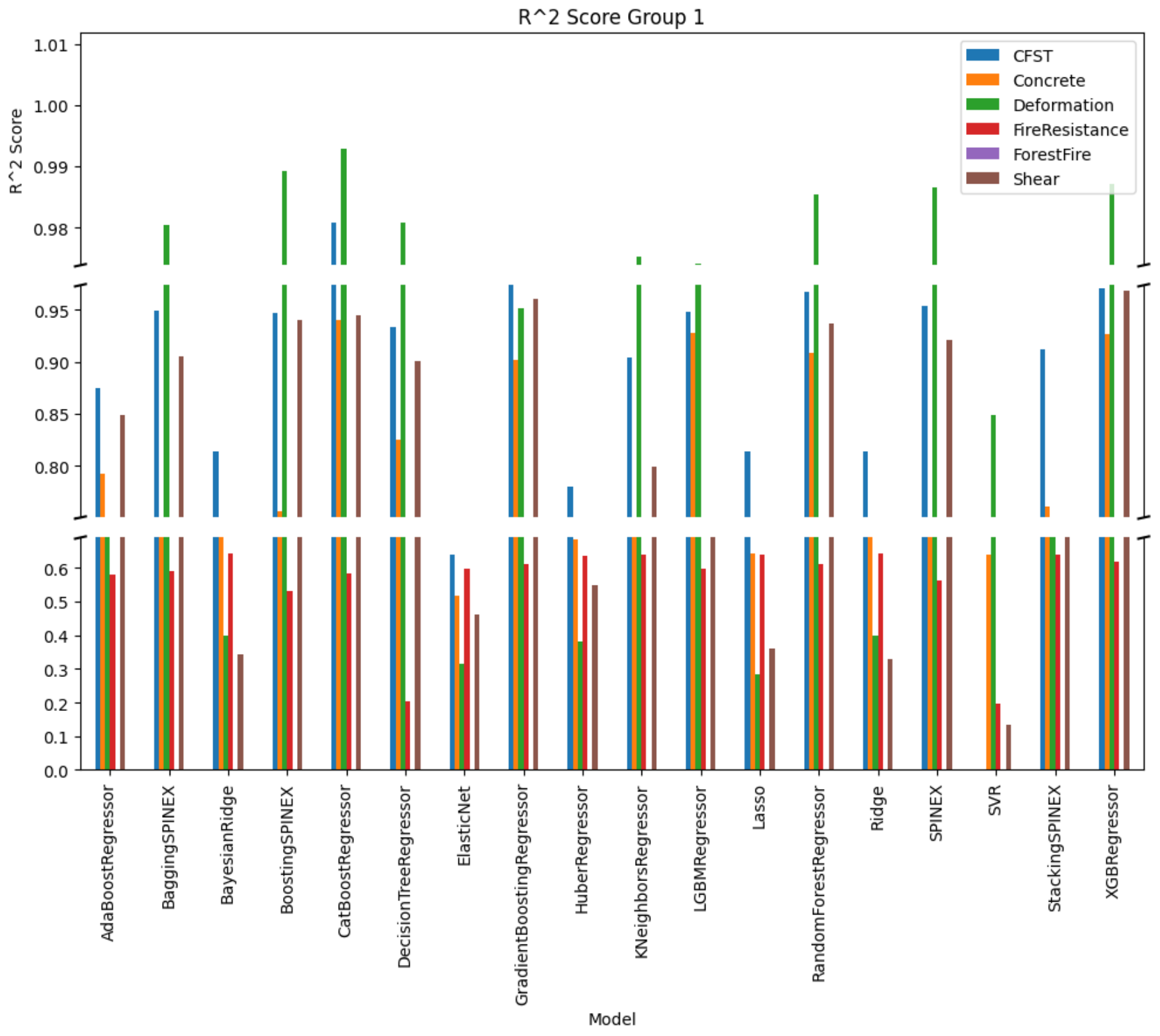

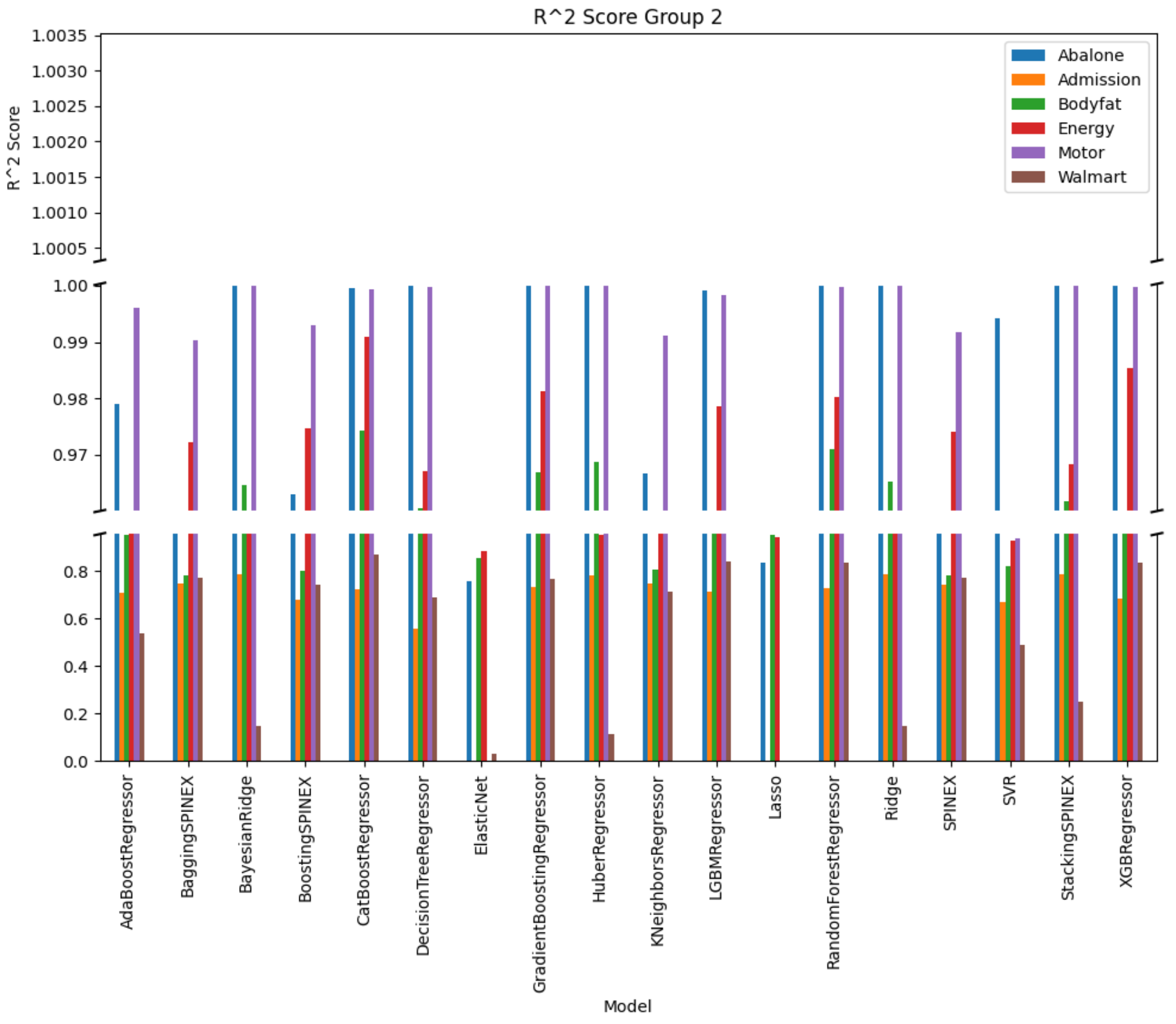

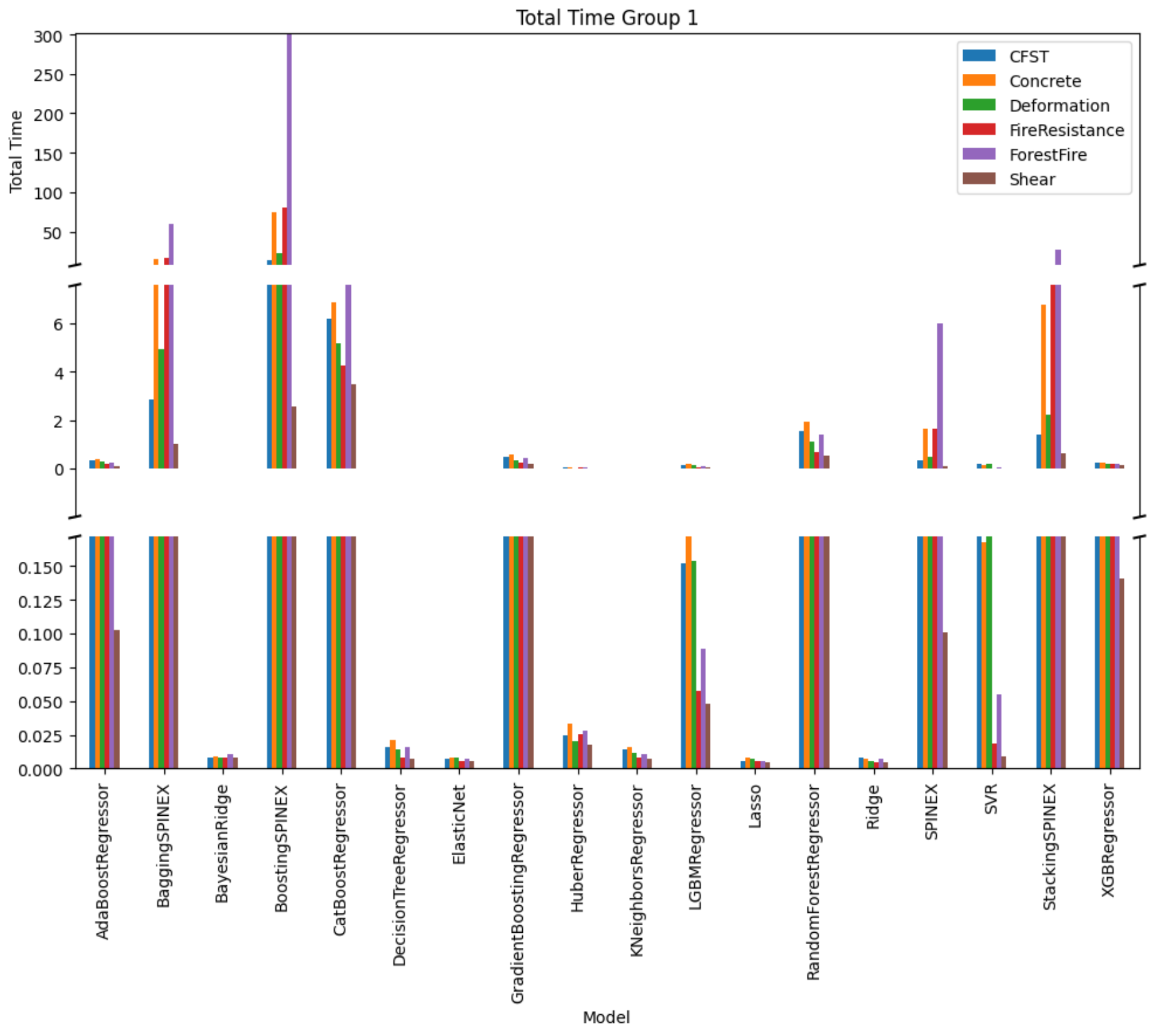

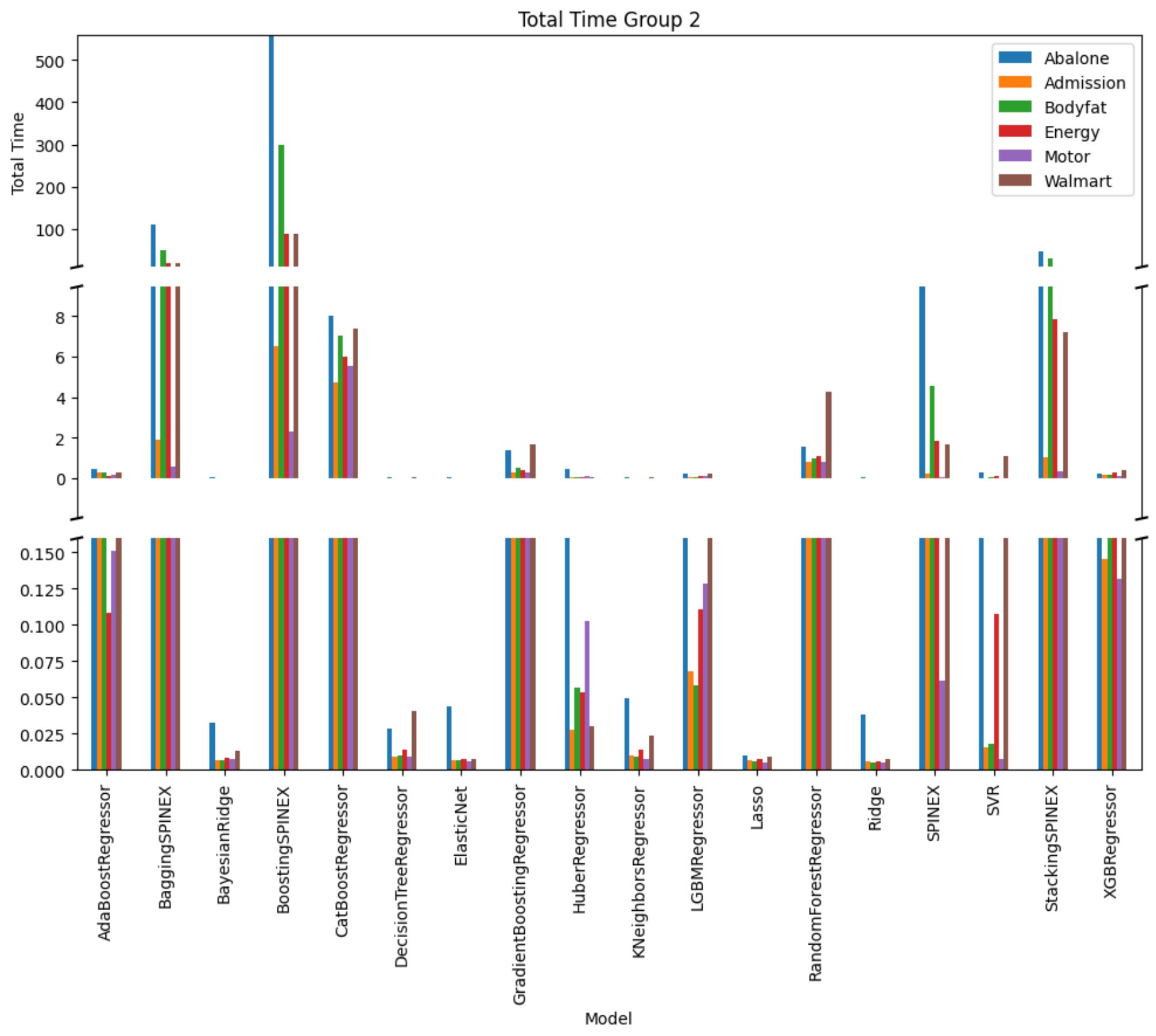

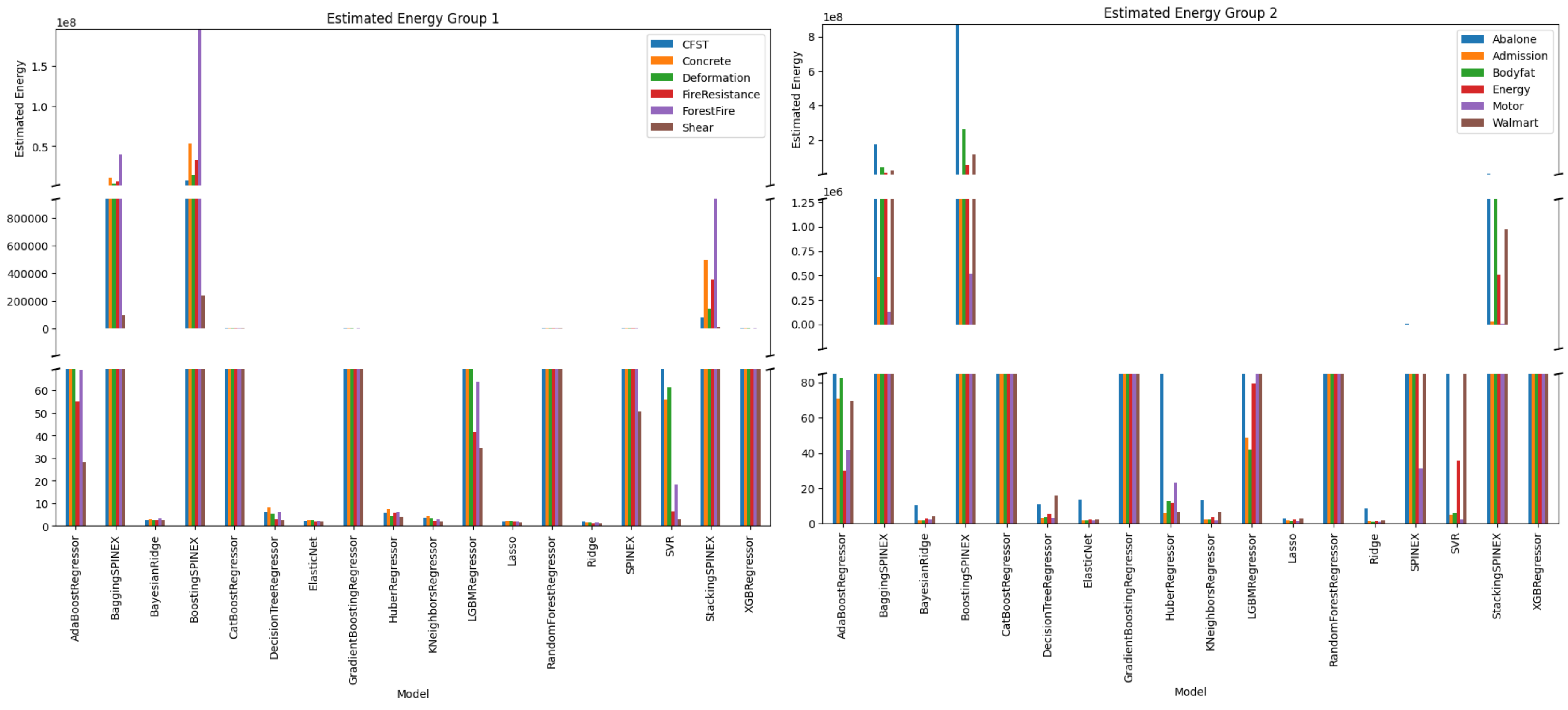

Fig. 4 Regression experiment on real data

### 3.2.2 Classification experiments

Here, 11 real datasets are used to examine SPINEX and the other ML models as a means for a second mean of validation. Table 9 lists details for each used dataset regarding the number of samples and features. Further information can be found in each dataset's reference.

Table 9 Datasets used in the classification analysis on real data

| Dataset | n_samples | n_features | Ref. |
|---|---|---|---|
| Fire-induced Spalling | 1062 | 16 | [42] |
| Pima Indians Diabetes | 768 | 8 | [43] |
| Bridge Failures | 299 | 7 | [44] |
| Concrete Condition in Situ | 9683 | 8 | [45] |
| Breast Cancer Wisconsin (Original) | 569 | 30 | [46,47] |
| Rice (Commeo and Osmancik) | 3810 | 7 | [48] |
| Bank Note Authentication | 1372 | 4 | [49] |
| Water Potability | 2011 | 9 | [50] |
| Machine Predictive Maintenance | 10000 | 5 | [51] |
| Depression Prediction | 1409 | 20 | [52] |
| Cars Purchase Decision | 1000 | 3 | [53] |

A look into Table 10 shows that GradientBoostingClassifier and CatBoostClassifier seem to rank constantly among the top two models in terms of accuracy. On the other hand, three versions of SPINEX (namely, StackingSPINEX and SPINEX) land at $6^{th}$ and $7^{th}$ in the overall ranking for accuracy. Figure 5 shows that despite the low ranking of SPINEX, this algorithm scored comparable performance to other traditional ML models, such as KNeighborsClassifier, DecisionTreeClassifier, LogisticRegression, and SVC.

Table 10 Ranking results of classification experiment on real data

| Models | Accuracy | LLE | AUC | Rank | Estimated Energy | Total Time | Rank |
|---|---|---|---|---|---|---|---|
| RandomForestClassifier | 3 | 5 | 3 | 1 | 9 | 13 | 9 |
| GradientBoostingClassifier | 1 | 2 | 2 | 2 | 8 | 12 | 7 |
| CatBoostClassifier | 2 | 1 | 1 | 3 | 10 | 15 | 10 |
| XGBClassifier | 5 | 4 | 5 | 3 | 6 | 8 | 5 |
| LGBMClassifier | 4 | 3 | 4 | 4 | 5 | 6 | 3 |
| AdaBoostClassifier | 6 | 11 | 6 | 5 | 4 | 10 | 5 |
| StackingSPINEX | 7 | 6 | 7 | 6 | 13 | 9 | 9 |
| SPINEX | 9 | 10 | 10 | 7 | 11 | 1 | 4 |
| DecisionTreeClassifier | 8 | 15 | 8 | 8 | 3 | 5 | 2 |
| BoostingSPINEX | 11 | 8 | 12 | 9 | 15 | 14 | 11 |
| LogisticRegression | 15 | 12 | 15 | 11 | 2 | 3 | 1 |
| BaggingSPINEX | 10 | 9 | 11 | 12 | 14 | 7 | 8 |
| KNeighborsClassifier | 13 | 13 | 9 | 12 | 1 | 4 | 1 |
| SPINEXClassifier(default) | 12 | 14 | 13 | 13 | 12 | 2 | 5 |
| SVC | 14 | 7 | 14 | 14 | 7 | 11 | 6 |

*SPINEX = (n_neighbors=20, distance_threshold=0.05, distance_threshold_decay=0.95, ensemble_method=None, metric='manhattan')

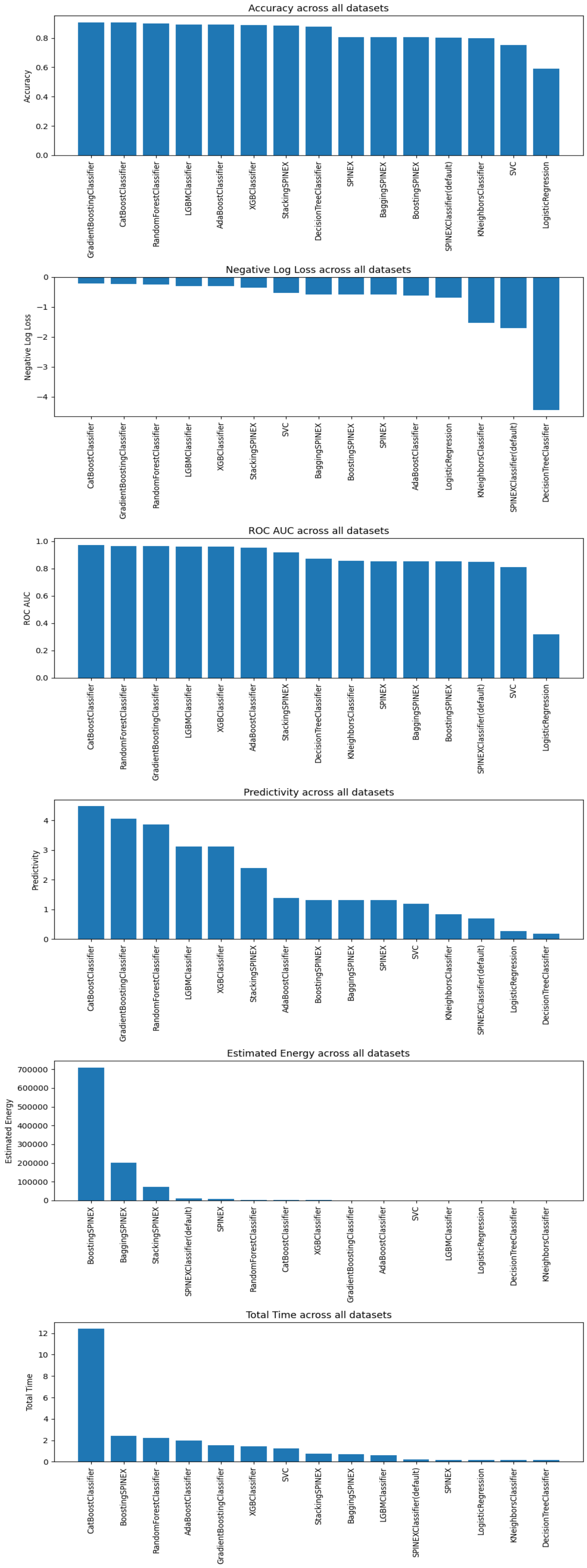

Fig. 4 Classification experiment on real data

### *3.3 Example with explainability*

Now, we show one example of the self-interpretability methods included within SPINEX. A sample case of synthetic dataset of 500 sample points and 5 features is presented herein. The results of the comparison in terms of accuracy and total time/energy used are shown in Fig. 5. As one can see, the SPINEX performance is well positioned against the other models.

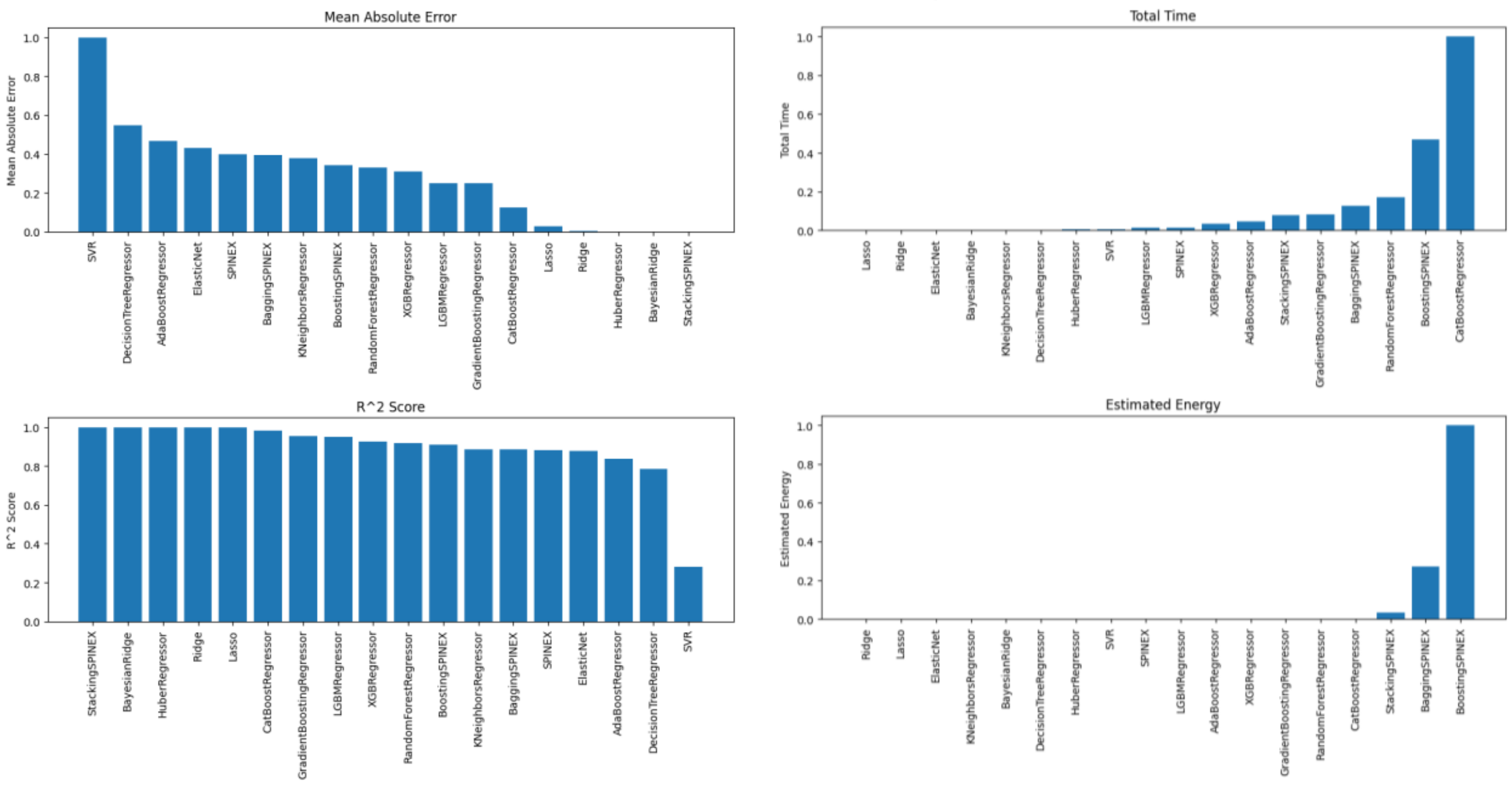

Fig. 5 Comparison of ML models

In terms of model explainability, Fig. 6 shows the calculated feature importance as described in Sec. 2.0. Comparatively speaking, the calculated trends of feature importance values seem to parallel that obtained from other models. The same figure also shows plots of other visualizations that can explain model behvaiour. Such visualization includes average interaction effects and feature combination between features. In addition, the pairwise interactions and feature importance at the global level and local level (for a particular instance) are also plotted – please refer to Sec. 2 for a detailed description of each of these visualizations. In addition, SPINEX can be accessed via: https://mznaser-clemson-spinexclassifier.streamlit.app/.

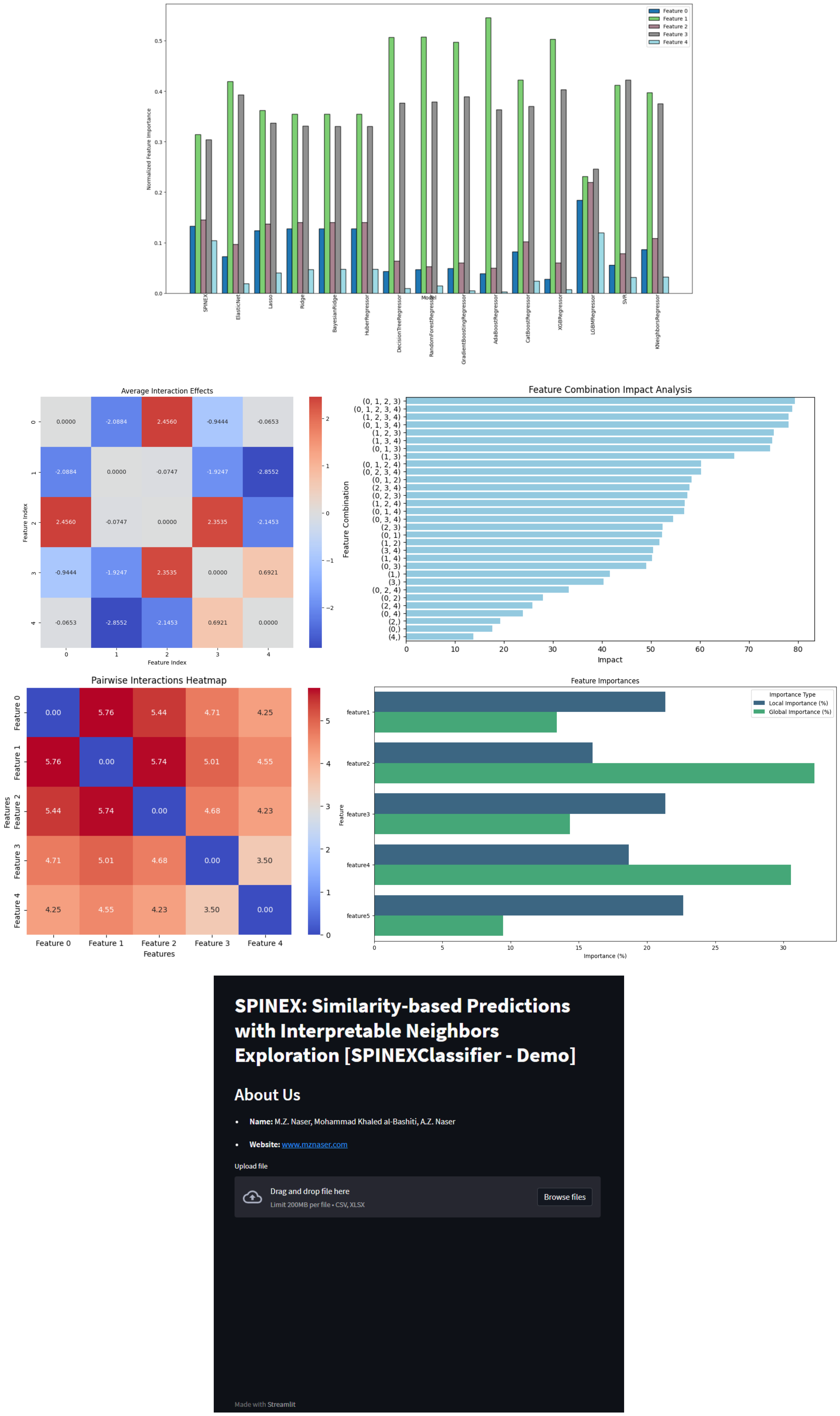

Fig. 6 Outcome of analysis and webapp

## 4.0 Conclusions

The SPINEX algorithm offers a novel approach for interpretable regression analysis by integrating ensemble learning with feature interaction analysis. It provides accurate predictions while unraveling the complex relationships between features and the target variable. The algorithm's neighbor-based feature importance and interaction effects offer transparent explanations for individual predictions, allowing users to gain insights into the model's decision-making process. It is expected that the performance of SPINEX will improve with further development efforts.

## Data Availability

Some or all data, models, or code that support the findings of this study are available from the corresponding author upon reasonable request.

## Conflict of Interest

The authors declare no conflict of interest.

## References

[1] J. Too, G. Liang, H. Chen, Memory-based Harris hawk optimization with learning agents: a feature selection approach, Eng. Comput. (2022). https://doi.org/10.1007/s00366-021-01479-4.

[2] I. Naruei, F. Keynia, Wild horse optimizer: a new meta-heuristic algorithm for solving engineering optimization problems, Eng. Comput. (2022). https://doi.org/10.1007/s00366-021-01438-z.

[3] C. Rudin, Stop explaining black box machine learning models for high stakes decisions and use interpretable models instead, Nat. Mach. Intell. (2019). https://doi.org/10.1038/s42256-019-0048-x.

[4] W.J. Murdoch, C. Singh, K. Kumbier, R. Abbasi-Asl, B. Yu, Definitions, methods, and applications in interpretable machine learning, Proc. Natl. Acad. Sci. U. S. A. (2019). https://doi.org/10.1073/pnas.1900654116.

[5] S.N. Van Der Veer, L. Riste, S. Cheraghi-Sohi, D.L. Phipps, M.P. Tully, K. Bozentko, S. Atwood, A. Hubbard, C. Wiper, M. Oswald, N. Peek, Trading off accuracy and explainability in AI decision-making: findings from 2 citizens' juries, J. Am. Med. Informatics Assoc. (2021). https://doi.org/10.1093/jamia/ocab127.

[6] H. Ding, I. Takigawa, H. Mamitsuka, S. Zhu, Similarity-basedmachine learning methods for predicting drug-target interactions: A brief review, Brief. Bioinform. (2013). https://doi.org/10.1093/bib/bbt056.

[7] G. Dudek, P. Pełka, Pattern similarity-based machine learning methods for mid-term load forecasting: A comparative study, Appl. Soft Comput. (2021). https://doi.org/10.1016/j.asoc.2021.107223.

[8] T. Hofmann, Learning the similarity of documents: An information-geometric approach to document retrieval and categorization, in: Adv. Neural Inf. Process. Syst., 2000.

[9] J. Wang, Y. Song, T. Leung, C. Rosenberg, J. Wang, J. Philbin, B. Chen, Y. Wu, Learning fine-grained image similarity with deep ranking, in: Proc. IEEE Comput. Soc. Conf. Comput. Vis. Pattern Recognit., 2014. https://doi.org/10.1109/CVPR.2014.180.

[10] B.M. Mathisen, A. Aamodt, K. Bach, H. Langseth, Learning similarity measures from data, Prog. Artif. Intell. (2020). https://doi.org/10.1007/s13748-019-00201-2.

[11] A. Charfi, S. Ammar Bouhamed, E. Bosse, I. Khanfir Kallel, W. Bouchaala, B. Solaiman, N. Derbel, Possibilistic Similarity Measures for Data Science and Machine Learning Applications, IEEE Access. (2020). https://doi.org/10.1109/ACCESS.2020.2979553.

[12] T. Widiyaningtyas, I. Hidayah, T.B. Adji, User profile correlation-based similarity (UPCSim) algorithm in movie recommendation system, J. Big Data. (2021). https://doi.org/10.1186/s40537-021-00425-x.

[13] F. Fkih, Similarity measures for Collaborative Filtering-based Recommender Systems: Review and experimental comparison, J. King Saud Univ. - Comput. Inf. Sci. (2021). https://doi.org/10.1016/j.jksuci.2021.09.014.

[14] P. Domingos, A few useful things to know about machine learning, Commun. ACM. (2012). https://doi.org/10.1145/2347736.2347755.

[15] K. Taunk, S. De, S. Verma, A. Swetapadma, A brief review of nearest neighbor algorithm for learning and classification, in: 2019 Int. Conf. Intell. Comput. Control Syst. ICCS 2019, 2019. https://doi.org/10.1109/ICCS45141.2019.9065747.

[16] S. Dhanabal, S. Chandramathi, A Review of various k-Nearest Neighbor Query Processing Techniques, Int. J. Comput. Appl. (2011).

[17] J. Laaksonen, E. Oja, Classification with learning k-nearest neighbors, in: IEEE Int. Conf. Neural Networks - Conf. Proc., 1996. https://doi.org/10.1109/icnn.1996.549118.

[18] D. Bera, R. Pratap, B.D. Verma, Dimensionality Reduction for Categorical Data, IEEE Trans. Knowl. Data Eng. (2021). https://doi.org/10.1109/TKDE.2021.3132373.

[19] E.Y. Boateng, J. Otoo, D.A. Abaye, Basic Tenets of Classification Algorithms K-Nearest-Neighbor, Support Vector Machine, Random Forest and Neural Network: A Review, J. Data Anal. Inf. Process. (2020). https://doi.org/10.4236/jdaip.2020.84020.

[20] H.A. Abu Alfeilat, A.B.A. Hassanat, O. Lasassmeh, A.S. Tarawneh, M.B. Alhasanat, H.S. Eyal Salman, V.B.S. Prasath, Effects of Distance Measure Choice on K-Nearest Neighbor Classifier Performance: A Review, Big Data. (2019). https://doi.org/10.1089/big.2018.0175.

[21] M. Loog, Nearest neighbor-based importance weighting, in: IEEE Int. Work. Mach. Learn. Signal Process. MLSP, 2012. https://doi.org/10.1109/MLSP.2012.6349714.

[22] M.Z. Naser, Do We Need Exotic Models? Engineering Metrics to Enable Green Machine Learning from Tackling Accuracy-Energy Trade-offs, J. Clean. Prod. 382 (2023) 135334. https://doi.org/10.1016/J.JCLEPRO.2022.135334.

[23] M.Z. Naser, V. Kodur, H.-T. Thai, R. Hawileh, J. Abdalla, V. V. Degtyarev, StructuresNet and FireNet: Benchmarking databases and machine learning algorithms in structural and fire engineering domains, J. Build. Eng. (2021) 102977. https://doi.org/10.1016/J.JOBE.2021.102977.

[24] M. van Smeden, K.G. Moons, J.A. de Groot, G.S. Collins, D.G. Altman, M.J. Eijkemans, J.B. Reitsma, Sample size for binary logistic prediction models: Beyond events per variable criteria:, Https://Doi.Org/10.1177/0962280218784726. 28 (2018) 2455–2474. https://doi.org/10.1177/0962280218784726.

[25] R.D. Riley, K.I.E. Snell, J. Ensor, D.L. Burke, F.E. Harrell, K.G.M. Moons, G.S. Collins, Minimum sample size for developing a multivariable prediction model: PART II - binary and time-to-event outcomes, Stat. Med. (2019). https://doi.org/10.1002/sim.7992.

[26] I. Frank, R. Todeschini, The data analysis handbook, 1994. https://books.google.com/books?hl=en&lr=&id=SXEpB0H6L3YC&oi=fnd&pg=PP1&ots=zfmIRO_XO5&sig=dSX6KJdkuav5zRNxaUdcftGSn2k (accessed June 21, 2019).

[27] F. Pedregosa, G. Varoquaux, A. Gramfort, V. Michel, B. Thirion, O. Grisel, M. Blondel, P. Prettenhofer, R. Weiss, V. Dubourg, J. Vanderplas, A. Passos, D. Cournapeau, M. Brucher, M. Perrot, E. Duchesnay, É. Duchesnay, E. Duchesnay, Scikit-learn: Machine learning in Python, J. Mach. Learn. Res. 12 (2011) 2825–2830.

[28] R. Kohavi, A study of cross-validation and bootstrap for accuracy estimation and model selection, Proc. 14th

Int. Jt. Conf. Artif. Intell. - Vol. 2. (1995).

[29] T.T. Wong, N.Y. Yang, Dependency Analysis of Accuracy Estimates in k-Fold Cross Validation, IEEE Trans. Knowl. Data Eng. (2017). https://doi.org/10.1109/TKDE.2017.2740926.

[30] M.Z. Naser, · Amir, H. Alavi, Error Metrics and Performance Fitness Indicators for Artificial Intelligence and Machine Learning in Engineering and Sciences, Archit. Struct. Constr. 2021. 1 (2021) 1–19. https://doi.org/10.1007/S44150-021-00015-8.

[31] A. Khare, Data for Admission in the University, Kaggle. (2022). https://www.kaggle.com/datasets/akshaydattatraykhare/data-for-admission-in-the-university.

[32] M.Z. Naser, V.K. Kodur, Explainable machine learning using real, synthetic and augmented fire tests to predict fire resistance and spalling of RC columns, Eng. Struct. 253 (2022) 113824. https://doi.org/10.1016/j.engstruct.2021.113824.

[33] I.-C.C. Yeh, Modeling of strength of high-performance concrete using artificial neural networks, Cem. Concr. Res. 28 (1998) 1797–1808. https://doi.org/10.1016/S0008-8846(98)00165-3.

[34] M.Z. Naser, AI-based cognitive framework for evaluating response of concrete structures in extreme conditions, Eng. Appl. Artif. Intell. 81 (2019) 437–449. https://www.sciencedirect.com/science/article/pii/S0952197619300466 (accessed April 1, 2019).

[35] S. Thai, H.-T. Thai, B. Uy, T. Ngo, M.Z. Naser, Test database on concrete-filled steel tubular columns, Mendeley, 2020. https://doi.org/10.17632/3XKNB3SDB5.5.

[36] U. Chowdhury, Energy Efficiency Data Set, Kaggel. (2022). https://www.kaggle.com/datasets/ujjwalchowdhury/energy-efficiency-data-set.

[37] Fedesoriano, Body Fat Prediction Dataset, Kaggle2. (2021). https://www.kaggle.com/datasets/fedesoriano/body-fat-prediction-dataset.

[38] P. Cortez, A. Morais, Forest Fires Data Set Portugal | Kaggle, (2007). https://www.kaggle.com/datasets/ishandutta/forest-fires-data-set-portugal (accessed July 11, 2022).

[39] Devphaib, Estimating the age of abalone at a seafood farm, Kaggle. (2022). https://www.kaggle.com/datasets/devzohaib/estimating-the-age-of-abalone-at-a-seafood-farm.

[40] Fedesoriano, Synchronous Machine Dataset, Kaggle. (2022). https://www.kaggle.com/datasets/fedesoriano/synchronous-machine-dataset.

[41] R. Patel, RETAIL ANALYSIS WITH WALMART SALES DATA, Kaggle. (2021). https://www.kaggle.com/datasets/rutuspatel/retail-analysis-with-walmart-sales-data.

[42] M.K. al-Bashiti, M.Z. Naser, Verifying domain knowledge and theories on Fire-induced spalling of concrete through eXplainable artificial intelligence, Constr. Build. Mater. 348 (2022) 128648. https://doi.org/10.1016/J.CONBUILDMAT.2022.128648.

[43] Pima Indians Diabetes Database, Kaggle. (2016). https://www.kaggle.com/datasets/uciml/pima-indians-diabetes-database.

[44] M. Abedi, M.Z. Naser, RAI: Rapid, Autonomous and Intelligent machine learning approach to identify fire-vulnerable bridges, Appl. Soft Comput. (2021). https://doi.org/10.1016/j.asoc.2021.107896.

[45] B.A. Young, A. Hall, L. Pilon, P. Gupta, G. Sant, Can the compressive strength of concrete be estimated from knowledge of the mixture proportions?: New insights from statistical analysis and machine learning methods, Cem. Concr. Res. 115 (2019) 379–388. https://doi.org/10.1016/j.cemconres.2018.09.006.

[46] W.H. Wolberg, O.L. Mangasarian, Multisurface method of pattern separation for medical diagnosis applied to breast cytology, Proc. Natl. Acad. Sci. U. S. A. (1990). https://doi.org/10.1073/pnas.87.23.9193.

[47] W. Wolberg, Breast Cancer Wisconsin (Original) Data Set, UCI Mach. Learn. Repos. (n.d.).

https://archive.ics.uci.edu/ml/datasets/Breast+Cancer+Wisconsin+%28Original%29.

[48] M. Koklu, Rice Dataset Commeo and Osmancik, Kaggle. (2022). https://www.kaggle.com/datasets/muratkokludataset/rice-dataset-commeo-and-osmancik.

[49] R. Saluja, Bank Note Authentication UCI data, Kaggle. (2018). https://www.kaggle.com/datasets/ritesaluja/bank-note-authentication-uci-data.

[50] A. Kadiwal, Water Quality, Kaggle. (2021). https://www.kaggle.com/datasets/adityakadiwal/water-potability.

[51] S. Bansal, Machine Predictive Maintenance, Kaggle. (2021). https://www.kaggle.com/datasets/shivamb/machine-predictive-maintenance-classification.

[52] D. Babativa, Depression dataset, Kaggle. (2023). https://www.kaggle.com/datasets/diegobabativa/depression.

[53] G. Santello, Cars - Purchase Decision Dataset, Kaggle. (2022). https://www.kaggle.com/datasets/gabrielsantello/cars-purchase-decision-dataset.

## Appendix

SPINEX and its derivatives will be provided herein.